\documentclass[letterpaper]{article} % DO NOT CHANGE THIS
\usepackage{aaai24}  % DO NOT CHANGE THIS
\usepackage{times}  % DO NOT CHANGE THIS
\usepackage{helvet}  % DO NOT CHANGE THIS
\usepackage{courier}  % DO NOT CHANGE THIS
\usepackage[hyphens]{url}  % DO NOT CHANGE THIS
\usepackage{graphicx} % DO NOT CHANGE THIS
\usepackage{natbib}  % DO NOT CHANGE THIS AND DO NOT ADD ANY OPTIONS TO IT
\usepackage{caption} % DO NOT CHANGE THIS AND DO NOT ADD ANY OPTIONS TO IT
\usepackage{algorithm}
\usepackage{newfloat}
\usepackage{listings}
\DeclareCaptionStyle{ruled}{labelfont=normalfont,labelsep=colon,strut=off} % DO NOT CHANGE THIS
\floatstyle{ruled}
\newfloat{listing}{tb}{lst}{}
\floatname{listing}{Listing}
\usepackage{xcolor}
\definecolor{yellow}{HTML}{ffd771}
\definecolor{blue}{HTML}{4473ba}

\usepackage{tikz}
\usetikzlibrary{patterns}

\usetikzlibrary{shapes.misc}
\tikzset{cross/.style={cross out, draw=black, minimum size=2*(#1-\pgflinewidth), inner sep=0pt, outer sep=0pt},
cross/.default={2pt}}

\usepackage[noend]{algpseudocode}
\usepackage{todonotes}
\usepackage{subcaption}
\usepackage{amsmath}
\usepackage{amsthm}
\usepackage{amssymb}
\usepackage{mathtools}
\usepackage{booktabs}
\usepackage{algorithm}

\usepackage{adjustbox}
\newtheorem{example}{Example}

\newtheorem{lemma}{Lemma}
\newtheorem{proposition}{Proposition}

\newtheorem{definition}{Definition}

\newcommand{\cinput}{\ensuremath{\mathbf{x}}}
\newcommand{\ccf}{\ensuremath{\mathbf{c}}}

\newcommand{\coutput}{\ensuremath{y}}
\newcommand{\classifier}{\ensuremath{\operatorname{Cl}}}
\newcommand{\dom}{\ensuremath{\mathcal{D}}}
\newcommand{\setdist}{\ensuremath{\operatorname{set-distance}}}
\newcommand{\cfd}{\ensuremath{\operatorname{cfd}}}

\title{Promoting Counterfactual Robustness through Diversity}
\author {
    Francesco Leofante\textsuperscript{\rm 1},
    Nico Potyka\textsuperscript{\rm 2}}
\affiliations {
    \textsuperscript{\rm 1} Department of Computing, Imperial College London, UK\\
    \textsuperscript{\rm 2} School of Computer Science and Informatics, Cardiff University, UK\\
    f.leofante@imperial.ac.uk, PotykaN@cardiff.ac.uk
}
\usepackage{bibentry}
\begin{document}

\maketitle

\begin{abstract}
Counterfactual explanations shed light on the decisions of black-box models by explaining
how an input can be altered to obtain a favourable decision from the model (e.g., when a loan application has been rejected).
However, as noted recently, counterfactual explainers may lack robustness in the sense that a minor change
in the input can cause a major change in the explanation. This can cause confusion on the user side and
open the door for adversarial attacks. In this paper, we study some sources of non-robustness. 
While there are fundamental reasons for why an explainer that returns a single counterfactual cannot be
robust in all instances, we show that some interesting robustness guarantees can be given by reporting 
multiple rather than a single counterfactual. Unfortunately, the number of counterfactuals that need to
be reported for the theoretical guarantees to hold can be prohibitively large. We therefore propose an approximation
algorithm that uses a diversity criterion to select a feasible number of most relevant explanations and study its robustness empirically. Our experiments indicate that our method improves the
state-of-the-art in generating robust explanations, while maintaining other desirable properties
and providing competitive computational performance.
\end{abstract}

\section{Introduction}

Counterfactual explanations support the outcome of a black-box machine learning 
model by explaining 
how the input could be changed to produce a different decision \cite{GuidottiMRTGP19,KarimiBSV23,StepinACP21,Wachter_17}.
Roughly speaking, a counterfactual explainer is
called robust, if a minor change in the input cannot cause a 
major change in the explanation. The actual change can be
quantified, for example, by the Euclidean distance, 
by the number of features
that have to be changed or by a cost associated with changing
the features. One motivation for robustness is user justifiability.
Two similar users would expect to get a similar explanation and 
an individual user may be surprised if a minor change in its
characteristics would result in a completely different explanation~\cite{Hancox-Li20}.
Robustness is also relevant from a fairness perspective
because some non-robust 
counterfactual explainers can be manipulated such that they
offer better explanations for particular subgroups \cite{SlackHLS21}.

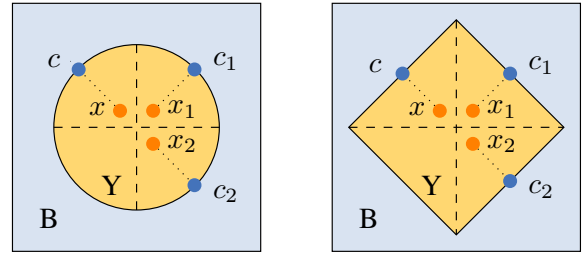
\begin{figure}[tb]
	\begin{subfigure}{0.45\columnwidth}
		\centering
		\scalebox{1.1}{\begin{tikzpicture}

% \draw[fill=yellow, fill opacity=0.5] (0,0) -- (0,2)  -- (1,2) -- (1.8,0) -- cycle ;

\draw[fill=blue, fill opacity=0.2] (0,0) -- (0,3)  -- (3,3) -- (3,0) -- cycle ;
\node at (0.4,0.4) {{ B}};

\draw[fill=yellow] (1.5,1.5) circle (1);
\node at (1.2,0.8) {{Y}};

\draw[dashed] (0.5,1.5) -- (2.5,1.5);
\draw[dashed] (1.5,0.5) -- (1.5,2.5);

% first cfx 
\draw[dotted] (1.3,1.7) -- (0.7,2.3);

\draw[orange,thick,fill=orange] (1.3,1.7) circle[radius=2pt];
\node[left] at (1.25,1.72) {{$x$}};

\draw[blue,thick,fill=blue] (0.8,2.2) circle[radius=2pt];
\node[left] at (0.7,2.3) {{$c$}};

% second cfx
\draw[dotted] (1.7,1.7) -- (2.2,2.2);

\draw[orange,thick,fill=orange] (1.7,1.7) circle[radius=2pt];
\node[right] at (1.75,1.72) {{$x_1$}};

\draw[blue,thick,fill=blue] (2.2,2.2) circle[radius=2pt];
\node[right] at (2.3,2.3) {{$c_1$}};

% third cfx
\draw[dotted] (1.7,1.3) -- (2.2,0.8);

\draw[orange,thick,fill=orange] (1.7,1.3) circle[radius=2pt];
\node[right] at (1.75,1.3) {{$x_2$}};

\draw[blue,thick,fill=blue] (2.2,0.8) circle[radius=2pt];
\node[right] at (2.3,0.7) {{$c_2$}};

\end{tikzpicture}}	
	\end{subfigure}
	\hfil
	\begin{subfigure}{0.45\columnwidth}
		\centering
  		\scalebox{1.1}{\begin{tikzpicture}

% \draw[fill=yellow, fill opacity=0.5] (0,0) -- (0,2)  -- (1,2) -- (1.8,0) -- cycle ;

\draw[fill=blue, fill opacity=0.2] (0,0) -- (0,3)  -- (3,3) -- (3,0) -- cycle ;
\node at (0.4,0.4) {{ B}};

\draw[fill=yellow] (0.2,1.5) -- (1.5,2.8)  -- (2.8,1.5) -- (1.5,0.2) -- cycle ;

\node at (1.2,0.8) {{Y}};

\draw[dashed] (0.2,1.5) -- (2.8,1.5);
\draw[dashed] (1.5,2.8) -- (1.5,0.2);

% first cfx 
\draw[dotted] (1.3,1.7) -- (0.85,2.15);

\draw[orange,thick,fill=orange] (1.3,1.7) circle[radius=2pt];
\node[left] at (1.25,1.72) {{$x$}};

\draw[blue,thick,fill=blue] (0.85,2.15) circle[radius=2pt];
\node[left] at (0.72,2.25) {{$c$}};

% second cfx
\draw[dotted] (1.7,1.7) -- (2.15,2.15);

\draw[orange,thick,fill=orange] (1.7,1.7) circle[radius=2pt];
\node[right] at (1.75,1.72) {{$x_1$}};

\draw[blue,thick,fill=blue] (2.15,2.15) circle[radius=2pt];
\node[right] at (2.25,2.25) {{$c_1$}};

% third cfx
\draw[dotted] (1.7,1.3) -- (2.15,0.85);

\draw[orange,thick,fill=orange] (1.7,1.3) circle[radius=2pt];
\node[right] at (1.75,1.3) {{$x_2$}};

\draw[blue,thick,fill=blue] (2.15,0.85) circle[radius=2pt];
\node[right] at (2.25,0.75) {{$c_2$}};

\end{tikzpicture}}	
	\end{subfigure}
	
	\caption{The figure shows the decision boundaries of two 
	classes $B$ (blue outer region) and $Y$ (yellow inner region) in two scenarios (left, right). In both cases, the points $x$, $x_1$, $x_2$ are relatively close, but their corresponding counterfactuals $\ccf, \ccf_1, \ccf_2$ on
	the decision boundary are relatively far away from each other
 wrt. Euclidean distance.}
	\label{fig:fund_rob_problem}
\end{figure}

There are different reasons for non-robustness. As noted in \cite{SlackHLS21},
local search methods as hill-climbing can be highly sensitive
to input perturbations and may therefore not be robust. All heuristic methods can be susceptible to such problems,
in particular if they rely on randomization. However, as
noted in \cite{fokkema2022attribution}, there are more 
fundamental problems that can cause robustness problems.
Intuitively, whenever an input is between two decision
boundaries on opposite sides, a minor change in the input
can result in a major change in the computed counterfactual.
We illustrate this in Figure \ref{fig:fund_rob_problem} for
two different decision boundaries and give additional 
explanations in the caption.
Note that this problem even applies to exact methods.
% \begin{figure}[tb]
% 	\centering
% 		\includegraphics[width=0.4\textwidth]{fig_problem.png}
% 	\caption{The figure shows the decision boundaries of two 
% 	classes $B$ (blue outer region) and $Y$ (yellow inner region) in two scenarios (left, right). In both cases, the points $x$, $x_1$, $x_2$ are relatively close, but their corresponding counterfactuals $\ccf, \ccf_1, \ccf_2$ on
% 	the decision boundary are relatively far away from each other
%  wrt. Euclidean distance.
% 	}
% 	\label{fig:fund_rob_problem}
% \end{figure}

The problem in Figure \ref{fig:fund_rob_problem} occurs if
the input $x$ is close to the center between decision
boundaries.
Clearly, a counterfactual explainer that returns a single counterfactual
is bound to lack robustness in such a scenario.
One intuitive idea to overcome this fundamental problem is to report multiple counterfactuals instead of a 
single one. 
In Section \ref{sec_cf_robustness}, we study this idea theoretically and identify some interesting 
cases under which an \emph{exhaustive explainer} 
can guarantee robustness by returning
all approximate counterfactuals with respect to
some tolerance parameter $\epsilon$.
Unfortunately, the exhaustive explainer is not practically
viable because the number of identified
counterfactuals can be prohibitively large or even infinite. The blowup is partially caused by sets of counterfactuals that are 
redundant in the sense that they are all very similar. Reporting all of them is not desirable from an explanation
perspective (because the explanation becomes too large) and is, indeed, not always necessary for guaranteeing 
robustness. 
% As identifying a minimal subset of non-redundant counterfactuals that guarantee robustness is computationally challenging, 
% we propose a simple approximation algorithm
To overcome these issues, we propose an approximation algorithm
in Section \ref{sec_cf_approx} that incrementally builds up a set of counterfactuals 
while filtering new candidates based on a diversity criterion. 
In Section \ref{sec_experiments}, we study the robustness and general performance of our approximation algorithm
and compare it to \texttt{DiCE}~\cite{mothilal2020explaining}, a state-of-the-art algorithm to generate sets of diverse explanations. Our experiments show that our algorithm is more robust and also outperforms \texttt{DiCE} along other metrics of interest, while maintaining superior runtime performance.

\section{Related Work}

% \todo[inline]{if the following is really relevant, 
% add it, otherwise ignore

% Some other approaches combine the two metrics instead and look for counterfactual explanations optimizing their combination. See for instance:

% @inproceedings{DBLP:conf/icml/ParmentierV21,
% author = {Axel Parmentier and
% Thibaut Vidal},
% editor = {Marina Meila and
% Tong Zhang},
% title = {Optimal Counterfactual Explanations in Tree Ensembles},
% booktitle = {Proceedings of the 38th International Conference on Machine Learning,
% {ICML} 2021, 18-24 July 2021, Virtual Event},
% series = {Proceedings of Machine Learning Research},
% volume = {139},
% pages = {8422--8431},
% publisher = {{PMLR}},
% year = {2021},
% url = {http://proceedings.mlr.press/v139/parmentier21a.html},
% timestamp = {Wed, 25 Aug 2021 17:11:17 +0200},
% biburl = {https://dblp.org/rec/conf/icml/ParmentierV21.bib},
% bibsource = {dblp computer science bibliography, https://dblp.org}
% }

% It would be interesting to compare your proposal with OCEAN, the algorithm implementing the approach described in DBLP:conf/icml/ParmentierV21. The comparison would be more significant than the one presented in the paper (with DICE and Proto), given that the latter approaches do not take plausibility into account.
% }
 
\textbf{Counterfactual explanations and their properties.} Several approaches have been proposed to compute counterfactual explanations for learning models.
The seminal work of Wachter et al~\cite{Wachter_17} used gradient-based optimisation to generate counterfactual explanations for neural networks. These counterfactuals are obtained by optimising a loss function that encourages their \emph{validity} (i.e., the counterfactual flips the classification outcome of the network) and \emph{proximity} (i.e., the counterfactual is close to the original input for which the explanation is sought under some distance metric). Following this initial proposal, other approaches have been developed to enforce additional properties on the explanations they produce. For instance, \texttt{DiCE}~\cite{mothilal2020explaining} proposed novel loss terms to generate sets of counterfactual explanations for a given input. By maximising the \emph{diversity} within the set, the authors provide a method to better approximate local decision boundaries of machine learning models, thus improving the explanatory power of counterfactuals. A different approach is proposed in \texttt{proto}~\cite{van2021interpretable}, where the authors present a method to generate counterfactual explanations that lie in the data manifold of the dataset in an attempt to improve their \emph{plausiblity}. The method relies on class prototypes identified by variational auto-encoders or kd-trees to guide the search for high-quality explanations. Moving away from continuous optimisation techniques for differentiable models,~\cite{MohammadiKBV21} cast the problem of finding counterfactual explanations as a constrained optimisation problem encoded and solved using Mixed-Integer Linear Programming. Similarly, \texttt{FACE}~\cite{karimi2020model} uses Satisfiability Modulo Theory solving to derive a model-agnostic counterfactual explanation algorithm. For further details on counterfactual explanations we refer to~\cite{KarimiBSV23}, which offers a recent survey on the state of the art in this area.

\textbf{Robustness and explainability.}
As explanations are increasingly used to guide decisions in areas with clear societal implications~\cite{COMPASS,HELOC}, their reliability has come under scrutiny. In particular, recent work has highlighted issues related to the \emph{robustness} of state-of-the-art counterfactual explainers. For instance,~\cite{upadhyay2021towards,BlackWF22,Jiang+23} study how the validity of counterfactual explanations is affected when the weights of a neural network are slightly altered, e.g., due to retraining or fine-tuning, observing that many state-of-the-art approaches fail to generate robust counterfactuals in this setting. Robustness to model changes is also investigated in~\cite{DuttaLMTM22}, where the authors consider tree-based classifiers and propose a statistical procedure to test the robustness of counterfactual explanations when minor modifications are applied to the tree. A related notion of robustness is also studied in~\cite{LeofanteBR23,pawelczyk2020counterfactual}, where the authors consider the more general setting of robustness under model multiplicity. 

In another line of work,~\cite{LeofanteLomuscio23a} show that the validity of a counterfactual explanation may be compromised by adversarial perturbations directly applied to the explanation itself. 
The authors discuss how such lack of \emph{model robustness}
(as opposed to \emph{input robustness} as we consider here)
hinder a transparent interaction between humans and AI agents. The authors propose to use formal verification techniques to counter this problem and derive a method to rigorously quantify the robustness of the explanations they produce. The same notion of robustness is also addressed in~\cite{Pawelczyketal23}, where a probabilistic method is proposed to generate robust counterfactuals. 

To the best of our knowledge, the only existing work that 
aims at improving input robustness is~\cite{SlackHLS21}. In particular, they showed that algorithms based on gradient search can be highly sensitive to changes in the input and may thus result in radically different explanations for very similar events. While they offer useful empirical observations on how to circumvent this robustness issue, they do not propose an algorithm to generate counterfactuals that are robust. In this paper we fill this gap and propose to move away from instance-based explanations and instead report multiple, diverse counterfactuals to improve the robustness of counterfactual explanations.

% some discussion about robustness problems (e.g., heuristic search, principal problems, workarounds)

% \todo[inline]{useful as a baseline and for inspiration}
% \cite{LooverenK21}
% \begin{enumerate}
%     \item It seems to me that the approach is heuristic in nature as it is based on minimizing a rather complicated loss function. They do not say much about the structure, but I would be very surprised if they could solve the problem exactly (they just say that they apply FISTA [2] to solve it). So, in particular, their approach may not be robust. It also relies on lots of hyperparameters that will probably affect the explanation. We could add it as a baseline and compare the robustness to our approach.
%     \item They propose two prototypical methods, one based on autoencoders, the other based on class-speficic k-d-trees. Only the latter seems to be related to what we do. The class-specific k-d-trees seem to be from [3] or [12]. If we can employ them here as well, I do not think that this is a problem because the application is different (for them, they are just a small part of the loss function, for us they may be a tool to identify good neighbors).
%     \item In general, they do not consider robustness at all and the ideas of
% 1) achieving robustness by partitioning the space and reporting multiple counterfactuals and
% 2) implementing the idea through diverse nearest neighbours
% go in a very different direction.
% \end{enumerate}
% \todo[inline]{end}

\section{Background}

We focus on classification problems over tabular data.
Our datasets are defined by a set of \emph{feature variables} (features)
$X_1, \dots, X_k$ and a \emph{class variable} $C$.
We let $D_1, \dots, D_k$ denote the \emph{domains} associated
with the variables and $L$ denote the \emph{class labels} associated
with the class $C$.
We let $\dom = \bigtimes_{i=1}^k D_i$ denote the set of all \emph{inputs (of the classification problem)}.
A domain $D$ is called 
% \emph{Boolean} if $D = \{0,1\}$, 
\emph{discrete} if $D$ is countable and 
\emph{continuous} if $D$ is an uncountable subset of $\mathbb{R}$.
A \emph{classification problem} $P = ((D_1, \dots, D_k), L, E)$ consists of domains, class labels and a set of training examples $E = \{(\cinput_i, \coutput_i) \mid 1 \leq i \leq N, \cinput_i \in \dom, \coutput_i \in L\}$. 
A training example $(\cinput_i, \coutput_i)$ consist of
an instantiation $\cinput_i$ of the variables and a class
label $\coutput_i$.
A \emph{classifier} is a function 
$\classifier: \dom \rightarrow L$
that assigns a class label  $l \in L$ to every input 
$\cinput$.

Counterfactual explanations explain how an input
can be changed to change the classification outcome.
For example, in a loan application scenario, users may be
interested in learning what they have to change in order to
be successful.
Formally, given a classifier $\classifier$ and
an input $\cinput \in \dom$ such that $\classifier(\cinput) = y$,
a \emph{counterfactual explanation} is an input
$\ccf \in \dom$ such that
$\classifier(\ccf) \neq y$ and
$\ccf$ is \emph{close to} $\cinput$. 
% \begin{itemize}
%     \item $\classifier(\ccf) \neq y$ and
%     \item $\ccf$ is \emph{close to} $\cinput$. 
% \end{itemize}
Proximity can be defined by different measures. This includes metrics like Euclidean or Manhattan distance, weighted variants and
measures that count the number of features that change.
Given one such distance measure $d: \dom \times \dom \rightarrow \mathbb{R}_0^+$, a point $\cinput' \in \dom$ may satisfy the
proximity constraint if it minimizes the distance among all
points that take a different class label or if the distance
is below a particular threshold.
Unless stated otherwise, we do
not make any assumptions about $d$
other than that it is non-negative.
Often, $d$ will be a metric,
that is, it will also satisfy
\begin{description}
    \item[Definiteness:] $d(\cinput_1,\cinput_2)=0$ if and only if $\cinput_1=\cinput_2$.
    \item[Symmetry:] $d(\cinput_1,\cinput_2)=d(\cinput_2,\cinput_1)$.
    \item[Triangle Inequality:] $d(\cinput_1,\cinput_3) \leq d(\cinput_1,\cinput_2) + d(\cinput_2,\cinput_3)$.
\end{description}
% If $\dom$ forms a vector space with zero vector $0$,
% one important special case are
% the norm-induced metrics. A norm is a function $\| . \|: \dom \rightarrow \mathbb{R}_0^+$ such that 
% the following properties are satisfied:
% \begin{description}
%     \item[Definiteness:] $\|x\|=0$ if and only if $x=0$.
%     \item[Absolute Homogeneity:] $\|s \cdot x\|= |s| \cdot \|x\|$ for all scalars $s$.
%     \item[Subadditivity:] $\|x + y\| \leq \|x\| + \|y\|$.
% \end{description}
% If $\| . \|$ is a norm, then the
%  $\| . \|$-induced distance
%  measure is $d_{\| . \|}(x,y) = \| x - y\|$ and it is well known
%  that $d_{\| . \|}$ is a metric. 
%  For example, the Euclidean, Manhattan and Chebyshev distances
%  are induced by the corresponding Euclidean, Manhattan and Chebyshev norms.

In the following, we will consider two types of 
counterfactuals.
\begin{definition}[Counterfactuals]
Given a reference point $\cinput \in \dom$
and a distance measure $d$, the \emph{counterfactual
distance (cfd) of $\cinput$} is defined as
\begin{equation}
\cfd(\cinput) = \inf \{d(\cinput, \cinput ') \mid 
\cinput' \in \dom, \classifier(\cinput) \neq \classifier(\cinput')\}.    
\end{equation}
A point $\ccf \in \dom$ such that $\classifier(c) \neq \classifier(\cinput)$ is called a
\emph{strong counterfactual (wrt. $\cinput$)} if
\begin{equation}
d(\cinput, \ccf) = \cfd(\cinput)    
\end{equation}
and, for $\epsilon \geq 0$, an \emph{$\epsilon$-approximate counterfactual (wrt. $\cinput$)} if
\begin{equation}
   d(\cinput, \ccf) \leq \cfd(\cinput) + \epsilon. 
\end{equation}
\end{definition}
Let us make some simple observations.
\begin{itemize}
    \item If the domain $\dom$ forms a complete metric space, 
    then for every boundary point $\cinput$ of a class
    such that $\cinput$ belongs to the class,
    we have $\cfd(\cinput)=0$. In this case, if the distance measure $d$ satisfies \emph{Definiteness}, there are no strong counterfactuals for $\cinput$ because the distance of a counterfactual must be non-zero.
    \item Every strong counterfactual is an $\epsilon$-approximate counterfactual for all $\epsilon \geq 0$.
\end{itemize}
In the following, we will consider approximate 
and exact counterfactual explainers.
Since our first observation implies that strong counterfactuals may not exist even though approximate
counterfactuals do, we allow that even an exact counterfactual explainer returns approximate 
counterfactuals in this boundary case. To this end,
we assume that exact explainers have a tolerance 
parameter $\epsilon$.
Let us note that, in practice, even explainers based
on exact optimization methods are typically only
$\epsilon$-approximate $d$-minimizing due to
limited precision.
\begin{definition}[Counterfactual Explainer]
\label{def_cf_explainer}
A \emph{counterfactual explainer} $E$ is a function that 
takes as input a classifier $\classifier$ and an input
$\cinput \in \dom$ and returns a set $E(\classifier, \cinput) \subset \dom$
of counterfactuals. We say that:
\begin{itemize}
    \item $E$ is
\emph{$\epsilon$-approximate $d$-minimizing}
if every $\ccf \in E(\classifier, \cinput)$
is an $\epsilon$-approximate counterfactual wrt. $\cinput$,
 \item $E$ is
\emph{$d$-minimizing (with tolerance $\epsilon > 0$)}
if every $\ccf \in E(\classifier, \cinput)$
is a strong counterfactual whenever $\cfd(\cinput)>0$
and an $\epsilon$-approximate counterfactual
if $\cfd(\cinput) = 0$.
\end{itemize}
\end{definition}

\section{Counterfactual Robustness Limitations and the Exhaustive Explainer $E^{\epsilon}_{exh}$}
\label{sec_cf_robustness}

Our goal is to design a counterfactual explainer 
that guarantees that whenever two inputs
$\cinput_1, \cinput_2 \in \dom$ are close, then their corresponding counterfactuals are close.
In general, we allow that 
counterfactual explainers return
a set of counterfactuals. 
There are various ways to 
measure distance between two sets.
Ideally, we would like to guarantee 
that for every 
counterfactual in $E(\classifier, \cinput_1)$,
there is a close counterfactual in
$E(\classifier, \cinput_2)$
and vice versa. To measure the extent to which this constraint is satisfied,
we consider two
set distance measures. The first one averages the distance of
counterfactuals between the sets,
whereas the second takes the maximum:
\begin{align}
\label{def_set_dist_avg}
    &\setdist^d_{\Sigma}(S_1,S_2) =  \frac{1}{2 \cdot \lvert S_1 \rvert} \sum_{\ccf_1 \in S_1} \min_{\ccf_2 \in S_2} d(\ccf_1, \ccf_2)
    \notag \\ 
     &\hspace{2.5cm}+ \frac{1}{2 \cdot \lvert S_2 \rvert} \sum_{\ccf_2 \in S_2} \min_{\ccf_1 \in S_1} d(\ccf_2, \ccf_1).
\end{align}
\begin{align}
\label{def_set_dist_max}
    &\setdist^d_{\max}(S_1,S_2) = \frac{1}{2} \big(  
    \max_{\ccf_1 \in S_1} \min_{\ccf_2 \in S_2} d(\ccf_1, \ccf_2) 
    \notag \\
     &\hspace{2.5cm}+\max_{\ccf_2 \in S_2} \min_{\ccf_1 \in S_1} d(\ccf_2, \ccf_1)
    \big).
\end{align}
The following lemma states some simple, but useful facts about our set distance measures.
\begin{lemma}
\begin{itemize}
    \item For all distance measures $d$, and all $\emptyset \subset S_1, S_2 \subseteq \dom$, 
    $\setdist^d_{\sum}(S_1,S_2) \leq \setdist^d_{\max}(S_1,S_2)$.
    \item If $d$ satisfies \emph{Symmetry}, $S_1 = \{\ccf_1\}$ and $S_2 = \{\ccf_2\}$ ($S_1$ and $S_2$ contain a single point),
    then $\setdist^d_{\sum}(S_1,S_2) = \setdist^d_{\max}(S_1,S_2) = d(\ccf_1,\ccf_2)$.
\end{itemize}    
\end{lemma}
\begin{proof}
See Appendix.
\end{proof}
The first item explains that the maximum distance is more conservative in the sense that it always returns a
distance at least as large as the sum-distance. That is, if we know that the maximum distance is smaller than some
$\epsilon > 0$, then so is the sum distance. The second item explains that the definitions generalize symmetric 
distance measures from points to sets. This is important for our experiments because it guarantees a fair comparison 
between counterfactual explainers that return a single counterfactual and those that return a set.

\begin{figure}[tb]
	\centering
		\includegraphics[width=\columnwidth]{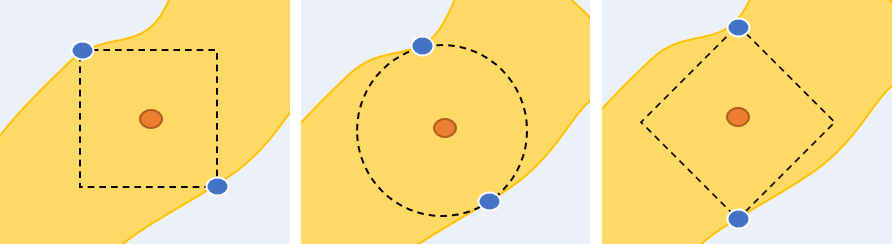}
	\caption{Critical $s$-balls with respect to Manhattan (left), Euclidean (middle) and
 Chebyshev (right) distance.
	}
	\label{fig:d_balls}
\end{figure}

We are now ready to give a first
formalization of robustness. 
Intuitively, we demand that if two inputs are close,
then the set-distance between their counterfactual explanations must be proportional to the distance
between the inputs. We use the
maximum distance in the definition since it is more conservative.
\begin{definition}[$(\epsilon,k)$-Robustness]
\label{def_e_k_robustness}
A counterfactual explainer $E$ is
\emph{$(\epsilon,k)$-robust} with respect to a distance
measure $d$ if for all inputs $\cinput_1, \cinput_2 \in \dom$
with $\classifier(\cinput_1) = \classifier(\cinput_2)$ and $d(\cinput_1, \cinput_2) < \epsilon$,
we have:
\begin{equation}
\label{def_k_robustness}
  \setdist^d_{\max}(E(\classifier, \cinput_1),E(\classifier, \cinput_2)) \leq k \cdot d(\cinput_1, \cinput_2).  
\end{equation}
\end{definition}
While $(\epsilon,k)$-robustness is desirable,
it may be impossible to satisfy.
The geometric intuition
is shown in Figure \ref{fig:fund_rob_problem}.
Whenever the input $\cinput$ 
is close to two counterfactuals
on opposite sides, a minor
change in $\cinput$ can cause
a major change in the corresponding 
counterfactual.
We make this intuition algebraically
more precise in the following example.
\begin{example}
\label{exp_euclidean_ball_cfs}
Consider a classification problem with $\dom = \mathbb{R}^n$
and two classes $Y$ (yellow region in Figure \ref{fig:fund_rob_problem}) and 
$B$ (blue region). We use the Euclidean distance as a distance measure and let $Y$ be the $r$-ball
$\{\cinput \in \mathbb{R}^n \mid
d(0,\cinput) \leq r\}$ centered
at $0$
as illustrated in Figure \ref{fig:fund_rob_problem}. 
Assume that the counterfactual
explainer finds counterfactuals
by minimizing the Euclidean distance, that is,
$E(\classifier, \cinput)
= \{\ccf \in \mathbb{R}^n \mid
d(\ccf,\cinput) = \cfd(\cinput)\}$.
Consider an input $\cinput \neq 0$
in the $r$-ball. Then there is a 
unique counterfactual $\ccf \in E(\classifier, \cinput)$. 
Now consider $\cinput_2 = - \cinput$. By symmetry,
the corresponding counterfactual
will be $\ccf_2 = -\ccf$. Furthermore, since
$\ccf, \ccf_2, 0$ are
collinear (they lie on the line
through $\ccf$ and $\ccf_2$),
we have 
$d(\ccf,\ccf_2) = d(\ccf,0) + 
d(0, \ccf_2) 
=2\cdot r$.
If we choose $\cinput, \cinput_2$
such that
$d(\cinput, \cinput_2) < \frac{2r}{k}$, then
$d(\cinput, \cinput_2) > k \cdot d(\cinput, \cinput_2)$.
Since we can choose $\cinput, \cinput_2$ such that the distance
is arbitrarily small, $E$ cannot
be $(\epsilon,k)$-robust for any choice of 
$\epsilon$ and $k$.
\end{example}
Note that the example can be generalized
to many other non-linear classification settings.
To illustrate this, 
Figure \ref{fig:d_balls} shows
some $s$-balls with respect to 
 Manhattan, Euclidean and
 Chebyshev distance. It should
 be geometrically clear that
 the same argument applies in these scenarios.
The essence of these examples is that the existence
of multiple counterfactuals can cause 
robustness problems. In order to avoid the problem,
we have to allow, at least, that counterfactual
explainers return more than one counterfactual.
However, even when returning multiple counterfactuals,
$(\epsilon,k)$-robustness cannot be guaranteed
because boundary counterfactuals
(i.e., counterfactuals with distance
$\cfd(\cinput) + \epsilon$)
are always lost
when making an arbitrarily small step away from them.
However, we can satisfy a weaker notion that
guarantees that strong counterfactuals are preserved.
\begin{definition}[Weak $\epsilon$-Robustness]
$E$ is
\emph{weakly $\epsilon$-robust} with respect to $d$ if for all inputs $\cinput_1, \cinput_2 \in \dom$
such that 
$\classifier(\cinput_1) = \classifier(\cinput_2)$
and $d(\cinput_1, \cinput_2) < \epsilon$,
we have that
% \begin{itemize}
%     \item 
    if $\cinput_{\ccf_1} \in E(\classifier, \cinput_1)$
    is a strong counterfactual with respect to $\cinput_1$,
    then  $\cinput_{\ccf_1} \in E(\classifier, \cinput_2)$ 
%     and
%     \item if $\cinput_{\ccf_2} \in E(\classifier, \cinput_2)$
%     is a strong counterfactual with respect to $\cinput_2$,
%     then $\cinput_{\ccf_2} \in E(\classifier, \cinput_1)$.
%     \todo[inline]{second condition may be redundant because we universally quantify anyway}
% \end{itemize}
\end{definition}
Conceptually, we can define a weakly $\frac{\epsilon}{2}$-robust 
counterfactual explainer by returning all $\epsilon$-approximate counterfactuals.
We call the corresponding explainer the \emph{exhaustive
$\epsilon$-approximate explainer} and
denote it by $E^{\epsilon}_{exh}$.
Before showing that $E^{\epsilon}_{exh}$ is indeed $\frac{\epsilon}{2}$-robust,
we state a simple lemma that is useful for proving 
robustness guarantees in general.
\begin{lemma}
If $d$ satisfies the triangle inequality, then
for all  $\cinput_1, \cinput_2 \in \dom$
such that 
$\classifier(\cinput_1) = \classifier(\cinput_2)$,
we have $\cfd(\cinput_1) \leq d(\cinput_1, \cinput_2) + \cfd(\cinput_2)$.
\end{lemma}
\begin{proof}
See Appendix.
\end{proof}
\begin{proposition}
If $d$ satisfies \emph{Symmetry} and
the \emph{Triangle Inequality}, then
$E^{\epsilon}_{exh}$ is weakly $\frac{\epsilon}{2}$-robust.    
\end{proposition}
\begin{proof}
See Appendix.
\end{proof}
We give a geometrical illustration of the proposition in the appendix after the proof.
The following example explains why we cannot give 
a general robustness guarantee for $\epsilon$-approximate counterfactuals that are not strong.
\begin{example}
\label{exp_boundary_approx_cfs}
Consider again the example in Figure \ref{fig:fund_rob_problem}.
Let $\epsilon > 0$ be minimally chosen such that 
$\ccf$ is an $\epsilon$-approximate counterfactual for $\cinput_2$. Then,
$\ccf \in E^{\epsilon}_{exh}(\classifier, \cinput_2)$.
However, when we let $\cinput_3$ be another input on the 
line segment between $\cinput_2$ and $\ccf_2$, then we will
lose $\ccf$ and all of its close neighbours in 
$E^{\epsilon}_{exh}(\classifier, \cinput_3)$.   
\end{example}
Let us note, however, that this case only occurs in boundary
cases. 
Intuitively, a counterfactual
is safer the closer it is to a strong counterfactual.
To make this intuition more precise, we define a notion of safety of counterfactuals based on a given input
that is to be explained and the tolerance parameter
$\epsilon$ of $E^{\epsilon}_{exh}$ (c.f., Def. \ref{def_cf_explainer}).
\begin{definition}
\label{label_def_safe_cf}
A counterfactual $\ccf$ of $\cinput$ is called 
\emph{$\delta$-safe with respect to 
$E^{\epsilon}_{exh}(\classifier, \cinput)$}
if $\delta = \cfd(\cinput) + \epsilon - d(\cinput, \ccf)$
and $\delta \geq 0$.
\end{definition}
For example, strong counterfactuals are
$\epsilon$-safe. Boundary $\epsilon$-approximate
counterfactuals as in Example 
\ref{exp_boundary_approx_cfs}
are $0$-safe. All other $\epsilon$-approximate counterfactuals are $\delta$-safe
for some $\delta \in (0, \epsilon)$.
As we show next, $E^{\epsilon}_{exh}$ also gives us some
robustness guarantees for $\delta$-safe counterfactuals
that are not strong.
Using our previous terminology, 
the following proposition roughly states that 
$E^{\epsilon}_{exh}$ is 
$\frac{\delta}{2}$-robust
for $\delta$-safe counterfactuals.
\begin{proposition}
\label{label_prop_safe_cf}
Suppose that $d$ satisfies \emph{Symmetry} and
the \emph{Triangle Inequality}.
If 
$\ccf \in E^{\epsilon}_{exh}(\classifier, \cinput)$
is $\delta$-safe, then for all
$\cinput' \in \dom$
such that
$\classifier(\cinput) = \classifier(\cinput')$
and
$d(\cinput, \cinput') < \frac{\delta}{2}$,
we have $\ccf \in E^{\epsilon}_{exh}(\classifier, \cinput')$.
\end{proposition}
\begin{proof}
See Appendix.
\end{proof}
Intuitively, $E^{\epsilon}_{exh}$ guarantees that
the closer a counterfactual is to being a strong counterfactual,
the further away we can move from the reference point
without losing the counterfactual.

At this point, we know that $E^{\epsilon}_{exh}$ gives us some interesting robustness guarantees.
However, $E^{\epsilon}_{exh}$ is not practical because
the number of $\epsilon$-approximate counterfactuals
is infinite in continuous domains and potentially exponentially large
(with respect to the number of features) in discrete domains.
In the next section, we will therefore focus on approximating
$E^{\epsilon}_{exh}$ by a subset of diverse counterfactuals
that provide a good tradeoff between explanation size
and robustness guarantees.

\section{Approximating $E^{\epsilon}_{exh}$}
\label{sec_cf_approx}

% We will now present an algorithm that approximates
% $E^{\epsilon}_{exh}$ by a subset of diverse counterfactuals.
% Our goal is to find a subset that is sufficiently small
% to be comprehensible while simultaneously being sufficiently
% large to guarantee robustness. 

\begin{figure*}[t!]
	\centering
	\begin{subfigure}{0.44\columnwidth}
		% \centering
		\scalebox{1.2}{\begin{tikzpicture}

\draw[fill=yellow, fill opacity=1] (0,0) -- (0,2)  -- (1,2) -- (1.8,0) -- cycle ;

\draw[fill=blue, fill opacity=0.2] (3,0) -- (3,2)  -- (1,2) -- (1.8,0) -- cycle ;

\draw[orange,thick, fill=orange] (1.1,0.4) circle[radius=2pt];

\draw[blue,thick, fill=blue] (1.9,0.3) circle[radius=2pt];
\node[below] at (1.9,0.3) {{\tiny \textbf{1}}};

\draw[blue,thick, fill=blue] (1.8,0.9) circle[radius=2pt];
\node[below] at (1.8,0.9) {{\tiny \textbf{2}}};

\draw[blue,thick, fill=blue] (2,1.1) circle[radius=2pt];
\node[below] at (2,1.1) {{\tiny \textbf{3}}};

\draw[blue,thick, fill=blue] (1.8,1.6) circle[radius=2pt];
\node[below] at (1.8,1.6) {{\tiny \textbf{4}}};

\draw[blue,thick, fill=blue] (2.8,1.8) circle[radius=2pt];
\node[below] at (2.8,1.8) {{\tiny \textbf{5}}};

\end{tikzpicture}}	
		\caption{}
		\label{fig:a}
	\end{subfigure}
	\hfil
	\begin{subfigure}{0.44\columnwidth}
		% \centering
  		\scalebox{1.2}{\begin{tikzpicture}

\draw[fill=yellow, fill opacity=1] (0,0) -- (0,2)  -- (1,2) -- (1.8,0) -- cycle ;

\draw[fill=blue, fill opacity=0.2] (3,0) -- (3,2)  -- (1,2) -- (1.8,0) -- cycle ;

\draw[orange,thick, fill=orange] (1.1,0.4) circle[radius=2pt];

\draw[blue,thick, fill=blue] (1.9,0.3) circle[radius=2pt];
\node[below] at (1.9,0.3) {{\tiny \textbf{1}}};

\draw[blue,thick, fill=blue] (1.8,0.9) circle[radius=2pt];
\node[below] at (1.8,0.9) {{\tiny \textbf{2}}};

\draw[blue,thick, fill=blue] (2,1.1) circle[radius=2pt];
\node[below] at (2,1.1) {{\tiny \textbf{3}}};

\draw[blue,thick, fill=blue] (1.8,1.6) circle[radius=2pt];
\node[below] at (1.8,1.6) {{\tiny \textbf{4}}};

\draw[gray,thick, fill=gray] (2.8,1.8) circle[radius=2pt];
\node[below,color=gray] at (2.8,1.8) {{\tiny \textbf{5}}};

\end{tikzpicture}}	
		\caption{}
		\label{fig:b}
	\end{subfigure}
	\hfil
	\begin{subfigure}{0.44\columnwidth}
		% \centering
		\scalebox{1.2}{\begin{tikzpicture}

\draw[fill=yellow, fill opacity=1] (0,0) -- (0,2)  -- (1,2) -- (1.8,0) -- cycle ;

\draw[fill=blue, fill opacity=0.2] (3,0) -- (3,2)  -- (1,2) -- (1.8,0) -- cycle ;

\draw[orange,thick, fill=orange] (1.1,0.4) circle[radius=2pt];

\draw[blue,thick, fill=blue] (1.9,0.3) circle[radius=2pt];
\node[below] at (1.9,0.3) {{\tiny \textbf{1}}};

\draw[blue,thick, fill=blue] (1.8,0.9) circle[radius=2pt];
\node[below] at (1.8,0.9) {{\tiny \textbf{2}}};

\draw[gray,thick, fill=gray] (2,1.1) circle[radius=2pt];
\node[below,color=gray] at (2,1.1) {{\tiny \textbf{3}}};

\draw[blue,thick, fill=blue] (1.8,1.6) circle[radius=2pt];
\node[below] at (1.8,1.6) {{\tiny \textbf{4}}};

\end{tikzpicture}}	
		\caption{}
		\label{fig:c}
	\end{subfigure}
	\hfil
	\begin{subfigure}{0.44\columnwidth}
		% \centering
		\scalebox{1.2}{\begin{tikzpicture}
\draw[fill=yellow, fill opacity=1] (0,0) -- (0,2)  -- (1,2) -- (1.8,0) -- cycle ;
\draw[fill=blue, fill opacity=0.2] (3,0) -- (3,2)  -- (1,2) -- (1.8,0) -- cycle ;
\draw[dotted] (1.1,0.4) -- (1.9,0.3);
\draw[dotted] (1.1,0.4) -- (1.8,0.9);
\draw[dotted] (1.1,0.4) -- (1.8,1.6);
\draw[orange,thick, fill=orange] (1.1,0.4) circle[radius=2pt];
\draw[blue,thick, fill=blue] (1.9,0.3) circle[radius=2pt];
\draw (1.67,0.33) node[cross=3pt,blue,thick] {};
\draw[blue,thick, fill=blue] (1.8,0.9) circle[radius=2pt];
\draw (1.53,0.7) node[cross=3pt,blue,thick] {};
\draw[blue,thick, fill=blue] (1.8,1.6) circle[radius=2pt];
\draw (1.42,0.95) node[cross=3pt,blue,thick] {};

% \draw[blue,thick, fill=blue] (1.9,0.3) circle[radius=2pt];
\node[below, color=blue, opacity=0.] at (1.9,0.3) {{\tiny \textbf{1}}};
\end{tikzpicture}}	
		\caption{}
		\label{fig:d}
	\end{subfigure}
	\caption{Pictorial representation of steps 1-4 performed by our approach.}
	\label{fig:algo_steps}
\end{figure*}
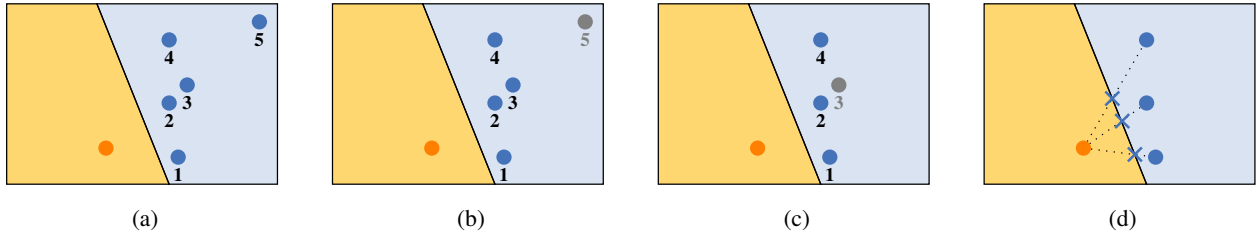

From a user perspective,
there is no point in reporting a large set of counterfactuals
that are all very similar. Instead, we should try to identify a
small set of diverse counterfactuals that
represent the counterfactuals 
in  $E^{\epsilon}_{exh}$ well. 
% Conceptually,
% this can be seen as a clustering problem, where we first 
% form clusters of similar counterfactuals and then report
% one representative counterfactual per cluster.
% While this idea is conceptually interesting, it is 
% computationally challenging. 
% We therefore resort to another
% approach.
We assume that our 
dataset $E$ is representative for the data that 
can occur in our domain and guide our search for
representative counterfactuals by
the examples that occur in $E$.
Before describing our approach in more detail, we give
a high-level overview of the four main steps it performs, which we also illustrate in Figure~\ref{fig:algo_steps}:

\begin{enumerate}
    \item Order examples based on distance from input (Fig~\ref{fig:a}).
    % \todo[inline]{if possible without too much effort, try alternative implementation based on sorting instead of k-d-trees}
    \item Filter examples based on distance (Fig~\ref{fig:b}).
    \item Filter remaining examples based on diversity (Fig~\ref{fig:c}).
    \item For every remaining example, compute a corresponding
    counterfactual via binary search (Fig~\ref{fig:d}).
\end{enumerate}
 
In step 1, we are given an input $\cinput$
with class $\classifier(\cinput) = l$.
We construct the set:
\begin{align}
  S_1 = 
  \{ (\cinput', d(\cinput, \cinput')) \mid
  (\cinput', y) \in E, \classifier(\cinput') \neq l
  \}
\end{align}
consisting of pairs of counterfactual points from $E$ and their distance to $\cinput$.
We order the elements in $S_1$
according to the distance in increasing order. 

In step 2, we restrict $S_1$ to the closest candidates. There are two natural options 
to cutoff candidates.
\begin{description}
    \item[Number-based:] pick the $M$ closest candidates.
    \item[Distance-based:] pick all candidates within a tolerance threshold.
\end{description}
The number-based approach cannot adapt to different characteristics
of $E$. For example, depending on whether the counterfactuals in $E$
are all very close (dense) or all very far away (sparse), we may want to make a different
choice of $M$.
The distance-based selection computes
the minimum distance of a counterfactual in $E$:
\begin{align}
\label{eq_def_m}
m = \min_{(\cinput',y') \in E} d(\cinput',\cinput),
\end{align}
and picks all candidates with distance at most $(1+\epsilon) \cdot m$
(all counterfactuals that are at most $100 \cdot \epsilon \%$ more distant than the closest counterfactual).
Hence, it will pick a larger number when $E$ is dense and a smaller 
number when $E$ is sparse.
We let $S_2$ denote the set obtained from $S_1$ by picking the closest candidates 
according to our selection criterion:
\begin{align}
S_2 = distanceFilter(S_1, \alpha),
\end{align}
where the parameter $\alpha$ determines the number of candidates (number-based)
or the tolerance threshold $\epsilon$ (distance-based).
% We can then construct a set $S_2$ of candidate counterfactuals by
% considering all points in $S_1$
% with distance
% at most $(1+\epsilon) \cdot m$. 
% That is, we let
% \begin{align}
% S_2 = \{
% (\cinput',d') \in S_1 \mid  d' \leq (1+\epsilon) \cdot m
% \},
% \end{align}
% where we keep the original order in $S_1$.

In step 3, we 
filter $S_2$ according to some
diversity criterion.
We consider two alternatives for this purpose.
\begin{description}
    \item[Angle-based:] We quantify the difference between two
    counterfactuals $\ccf_1, \ccf_2$ by the angle between them relative to the input $\cinput$. Formally, we compute the
    cosine distance between $(\cinput_1 - \cinput)$ and $(\cinput_2 - \cinput)$.
    \item[Distance-based:] Based on the distance between $\ccf_1$ and $\ccf_2$.
\end{description}
We create the filtered set $S_3$ in a greedy fashion.
Starting with the set containing only the closest counterfactual,
we successively add
elements from $S_2$ (still ordered by distance) if they are sufficiently different from all
candidates that have already been
added. For the cosine distance,
we use the intuition that the 
angle between the vectors should be sufficiently large. For the distance-based
filter we demand that the distance
between two counterfactuals is at least $100 \cdot \beta \%$ more than $m$ (Equation \ref{eq_def_m}) for some $\beta \in [0,1]$, which leads to the constraint 
$d(x,y) \geq (1 + \beta) \cdot m$.
We define $S_3$ by
filtering $S_2$ based on the angle or distance.
\begin{align}
    S_3 = diversityFilter(S_2, \beta),
\end{align}
where the parameter $\beta$ determines the angular (angle-based) or distance (distance-based) threshold.

In step 4, we compute 
counterfactuals from the remaining
candidates in $S_3$. To this end, 
we perform a binary search 
(Algorithm~\ref{alg:algo1})
for 
every $\ccf \in S_3$ to find the 
closest counterfactual to $\cinput$
on the line segment between $\cinput$ and $\ccf$.
Our algorithm finally returns:
\begin{align}
 C = binarySearch(S_3, \cinput).   
\end{align}
% \begin{figure}
% \begin{align*}
%     &\textbf{Input: } \textit{ reference point } \cinput, \textit{ counterfactual } \ccf, \textit{ desired accuracy } \gamma \\[0.05cm]
%     &\textbf{Output:} \textit{ closest counterfactual on line segment between $\cinput$ and $\ccf$} \\[0.3cm]
%     % 
%     &\textbf{WHILE}(d(\cinput, \ccf) > \gamma): \\[0.05cm]
%     &\hspace{0.6cm} \cinput' \leftarrow \frac{\cinput + \ccf}{2} \\[0.05cm] 
%     &\hspace{0.6cm} \textbf{IF } \classifier(\cinput') = \classifier(\cinput) : \\[0.05cm] 
%     &\hspace{0.6cm}\hspace{0.6cm} \cinput \leftarrow \cinput' \\[0.05cm] 
%     &\hspace{0.6cm} \textbf{ELSE }\\[0.05cm] 
%     &\hspace{0.6cm}\hspace{0.6cm} \ccf \leftarrow \cinput' \\[0.05cm] 
%     &\textbf{RETURN } \ccf
% \end{align*}
%     \caption{Binary line-search algorithm.}
%     \label{fig:search_algorithm}
% \end{figure}
\algrenewcommand\algorithmicrequire{\textbf{Input:}}
\algrenewcommand\algorithmicensure{\textbf{Output:}}
\begin{algorithm}[t!]
% \setstretch{1.25}
\caption{Binary linear-search algorithm} \label{alg:algo1}
\begin{algorithmic}[1]
\Require reference point $\cinput$, counterfactual $\ccf$, desired accuracy $\gamma$
\Ensure closest counterfactual on line segment between $\cinput$ and $\ccf$
\While{$d(\cinput, \ccf) > \gamma$}
\State $\cinput' \leftarrow \frac{\cinput + \ccf}{2}$
\If{$\classifier(\cinput') = \classifier(\cinput)$}
\State $\cinput \leftarrow \cinput'$
\Else
\State $\ccf \leftarrow \cinput'$
\EndIf
\EndWhile
\State \textbf{return} $\ccf$
\end{algorithmic}
\end{algorithm}

We give a short runtime analysis of our algorithm in the following proposition.
\begin{proposition}
Consider a classification problem with $k$ features and $N$ examples.
\begin{itemize}
    \item Step 1 can be computed in time $O(N \cdot (k + \log N))$.
    \item Step 2 can be computed in time $O(N \cdot k)$.
    \item Let $M$ be the number of points remaining after step 2. Assuming that the distance can be
    computed in time $O(k)$, Step 3 can be computed in time
     $O(M^2 \cdot k)$.
     \item Let $T_{\classifier}(k)$ be the runtime function of the classifier and let 
     $D_{max}$ be the maximum distance between the reference point $\cinput$ and one of the remaining
     points. 
     Step 4 can be computed in time $O(M \cdot (k + T_{\classifier}(k)) \cdot \log_2 \frac{D_{max}}{\gamma})$.
\end{itemize}    
\end{proposition}
\begin{proof}
See Appendix. 
\end{proof}
Let us note that many classifiers can classify examples in linear time, that is, $T_{\classifier}(k) = O(k)$. The overall runtime is then roughly quadratic with respect
to the number of features and the number of candidates remaining after step 2 and log-linear in the number of all examples.

We implemented a first prototype of our algorithm\footnote{Available at: \url{https://github.com/fraleo/robust_counterfactuals_aaai24}}.
Instead of sorting and maintaining the points manually, we use a k-d-tree. This is likely to increase runtime, but simplified the implementation.
% \todo[inline]{to make the choices of options less ad-hoc, 
% it may be good to make
% an experimental comparison between the different options
% (naive grid search?)
% and to try some different hyperparameter settings. The results could be
% put in the appendix and referred to for motivation of our settings}

\section{Experimental Analysis}
\label{sec_experiments}

% \todo[inline]{Rev 3: 
% about using training set E. I'm puzzled that no test set is used. Is a classifier CL given independently from E? This is unclear in the proposed setting}
% \todo[inline]{All Reviewers: more datasets, more features. More information about hyperparameter 
% setting (we should perhaps do some hyperparameter tuning first :D).

% choice of what is the parameter setup undefined and unclear how to optimise the parameters, e.g., number-based, distance-based, angle-based, distance-based, $\alpha, \beta,\delta, \gamma, \epsilon$, Unclear what you did use in the experiments. Also, many more datasets are needed than those presented here, diverse in instances size, number and type of features to make any significant comparison with other methods. Also, why did you stick on $L_1$ distance and restrict to only 5 counterfactuals?}
% \todo[inline]{R2: In some cases the explainer proposed in this work does not give the best result, in comparison to the other ones. How would the author(s) explain the drawback in general?}

In the previous sections we laid the theoretical foundations for a framework to generate counterfactual explanations that are robust to changes in the input. We then introduced an approximation algorithm that uses diversity to generate robust counterfactual explanations while maintaining computational feasibility. In this section, we evaluate the performance of our approximation algorithm empirically. To this end, we compare our method with \texttt{DiCE}, the de-facto standard approach to generate sets of diverse counterfactual explanations. We study the robustness of the two methods along different metrics and show that our approach outperforms \texttt{DiCE} in most cases.
%     \item in \S~\ref{sec:single_exp} we analyse the performance of our approach when used to generate single counterfactual explanations. To this end, we compare against \texttt{DiCE} and \texttt{proto}. Our experiments show that our approach is competitive even in this setting, despite not being specifically designed to generate single explanations. 
% \end{itemize}

\subsection{Experimental setup}
\label{sec:setup}

\textbf{Datasets.} We consider five binary classification datasets commonly used in the literature: \textit{diabetes}~\cite{smith1988using}, \textit{no2}~\cite{OpenML2013}, \textit{credit}~\cite{Dua2019}, \textit{spambase}~\cite{misc_spambase_94} and \textit{online news popularity}~\cite{misc_online_news_popularity_332}. Our selection includes both low- and high-dimensionality data, which allows to evaluate the applicability of our approach in both scenarios. We split each dataset into a training set and test set; more details about the datasets can be found in the appendix.
 
\textbf{Models and algorithms.} We train neural network classifier with two hidden layers ($20$ and $10$ neurons respectively) for each dataset and use two algorithms to generate diverse explanations: ours and \texttt{DiCE}\footnote{Available at: \url{https://github.com/interpretml/DiCE}}~\cite{mothilal2020explaining}, which uses gradient-based optimisation to generate sets of explanations under a loss function that optimises their diversity and proximity to the input. 

% \todo[inline]{Comment on choice of parameters}
\textbf{Hyperparameters.} \texttt{DiCE} is run with default parameters from the respective library. As for our approach, the following configuration is used: number-based selection in Step 2 with $\alpha = 50$ for \textit{diabetes} and \textit{no2}, and $\alpha = 1000$ for the remaining datasets; angle-based filter with $\beta = 0.5$ in Step 3 and $\gamma = 0.1$ in Step 4. More details on hyperparameter selection can be found in the appendix.

\textbf{Protocol.} Counterfactual explanations are generated following the same protocol. Given an input $\cinput$ a set of counterfactual explanations is generated using one among the algorithms considered. Then, a Gaussian distribution centered at $\cinput$ is sampled to obtain a new input $\cinput_2$ of the same class. This input is then used to test the robustness of the counterfactual explainers as follows. The same counterfactual explanation algorithm is run on $\cinput_2$, the resulting set of counterfactual explanations is evaluated along different metrics that we describe in the next section. We run this protocol three times for each input $\cinput$ and collect average and standard deviation for each of the metrics we consider. We use this protocol to evaluate experimentally to which extent our approximation can maintain the theoretical robustness guarantees of $E^{\epsilon}_{exh}$.

\textbf{Hardware.} All experiments were conducted on standard PC running Ubuntu 20.04.6 LTS, with 15GB RAM and processor Intel(R) Core(TM) i7-8700 CPU @ 3.20GHz.

\begin{table*}[t!]
    \centering
    \resizebox{1\textwidth}{!}{
    \begin{tabular}{ccccc|cccc|cccc}
        \cline{2-13}&
        \multicolumn{4}{c}{\textbf{diabetes}} &
        \multicolumn{4}{c}{\textbf{no2}} &
        \multicolumn{4}{c}{\textbf{news}} \\
        \cline{2-13}
        & ours (L1) & \texttt{DiCE} (L1)  & ours (L2) & \texttt{DiCE} (L2) & ours (L1) & \texttt{DiCE} (L1) & ours (L2) & \texttt{DiCE} (L2) & ours (L1) & \texttt{DiCE} (L1) & ours (L2) & \texttt{DiCE} (L2)\\
        \hline
         validity & 100\% & 100\% & 100\% & 100\% & 100\% & 100\% & 100\% & 100\% & 100\% & 100\% & 100\% & 100\%  \\ \hline
         
        $k$-distance  & 1.13 $\pm$ 0.43 & 1.83 $\pm$ 0.35 &0.52 $\pm$ 0.20 & 1.06 $\pm$ 0.18 & 
                        0.62 $\pm$0.23 & 1.26 $\pm$ 0.22 & 0.31 $\pm$ 0.11 & 0.80 $\pm$ 0.15 & 
                        2.70 $\pm$ 0.97 & 3.59 $\pm$ 0.78 & 0.75 $\pm$ 0.27 & 1.47 $\pm$ 0.24  \\ \hline
        
        $k$-diversity  & 1.39 $\pm$ 0.46 & 1.32 $\pm$ 0.19 & 0.63 $\pm$ 0.22 & 0.77 $\pm$ 0.10 & 
                       0.78 $\pm$ 0.28 & 0.87 $\pm$ 0.16 & 0.30 $\pm$ 0.14 & 0.50 $\pm$ 0.09 &
                       3.45 $\pm$ 1.24 & 1.91 $\pm$ 0.93 & 0.94 $\pm$ 0.35 & 0.78 $\pm$ 0.40 \\ \hline
        
        $\setdist^d_{\Sigma}$ & 0.21 $\pm$ 0.20 & 0.33 $\pm$0.1 & 0.09 $\pm$ 0.03 & 0.18 $\pm$ 0.06 &
                                0.16 $\pm$ 0.12 & 0.87 $\pm$ 0.16 & 0.07 $\pm$ 0.05 & 0.16 $\pm$ 0.07 &
                                0.88 $\pm$ 0.78 & 1.73 $\pm$ 0.72 & 0.21 $\pm$ 0.21 & 0.61 $\pm$ 0.30\\ \hline
        
       $\setdist^d_{max}$  &  0.51 $\pm$ 0.44 &  0.66 $\pm$ 0.25 & 0.24 $\pm$ 0.20 & 0.38 $\pm$ 0.16 &
                              0.33 $\pm$ 0.24 & 0.47 $\pm$ 0.24 & 0.16 $\pm$ 0.11 & 0.26 $\pm$ 0.13 &
                              1.94 $\pm$ 1.41 & 2.59 $\pm$ 1.00 & 0.52 $\pm$ 0.44 & 0.94 $\pm$ 0.40 \\ \hline
       
       Time (s) & 0.02 $\pm$ 0.00 & 72.66 $\pm$ 32.57 & 0.02 $\pm$ 0.00 & 75.22 $\pm$ 35.29 &
                0.02 $\pm$ 0.01 & 130.78 $\pm$ 13.945 & 0.02 $\pm$ 0.01 & 129.13 $\pm$ 134. 66 &
                0.30 $\pm$ 0.08 & 338.89 $\pm$ 11.59 & 0.30 $\pm$ 0.00 & 340.26 $\pm$ 117.52\\
       \hline
    \end{tabular}
    }
    \caption{Comparison between \texttt{DiCE} and our method (angle-based) on $50$ instances for $\beta = 0.5$ and $\gamma = 0.1$ We report average and standard deviation for each metric (validity excluded). 
    % Best results for each metric highlighted in bold.
    }
    \label{tab:final-results-angle-min}
\end{table*}
\begin{table*}[t!]
    \centering
    \resizebox{1\textwidth}{!}{
    \begin{tabular}{ccccc|cccc|cccc}
        \cline{2-13}&
        \multicolumn{4}{c}{\textbf{diabetes}} &
        \multicolumn{4}{c}{\textbf{no2}} &
        \multicolumn{4}{c}{\textbf{news}} \\
        \cline{2-13}
        & ours (L1) & \texttt{DiCE} (L1)  & ours (L2) & \texttt{DiCE} (L2) & ours (L1) & \texttt{DiCE} (L1) & ours (L2) & \texttt{DiCE} (L2) & ours (L1) & \texttt{DiCE} (L1) & ours (L2) & \texttt{DiCE} (L2)\\
        \hline
         validity & 100\% & 100\% & 100\% & 100\% & 100\% & 100\% & 100\% & 100\%  & 100\% & 100\% & 100\% & 100\%  \\ \hline
         
        $k$-distance  & 1.38 $\pm$ 0.29 & 1.83 $\pm$ 0.35 & 0.64 $\pm$ 0.14 & 1.06 $\pm$ 0.18 & 
                        0.87 $\pm$ 0.20 & 1.26 $\pm$ 0.22 & 0.43 $\pm$ 0.09 & 0.80 $\pm$ 0.15 & 
                        3.53 $\pm$ 0.90 & 3.59 $\pm$ 0.78 & 0.98 $\pm$ 0.26 & 1.47 $\pm$ 0.24  \\ \hline
        
        $k$-diversity  & 1.71 $\pm$ 0.3 & 1.32 $\pm$ 0.19 & 0.78 $\pm$ 0.15 & 0.77 $\pm$ 0.10 & 
                       1.11 $\pm$ 0.24 & 0.87 $\pm$ 0.16 & 0.55 $\pm$ 0.10 & 0.50 $\pm$ 0.09 &
                       4.35 $\pm$ 1.17 & 1.91 $\pm$ 0.93 & 1.20 $\pm$ 0.34 & 0.78 $\pm$ 0.40 \\ \hline
        
        $\setdist^d_{\Sigma}$ & 0.22 $\pm$ 0.24 & 0.33 $\pm$0.1 & 0.10 $\pm$ 0.11 & 0.18 $\pm$ 0.06 &
                                0.15 $\pm$ 0.16 & 0.87 $\pm$ 0.16 & 0.07 $\pm$ 0.08 & 0.16 $\pm$ 0.07 &
                                0.79 $\pm$ 0.97 & 1.73 $\pm$ 0.72 & 0.21 $\pm$ 0.27 & 0.61 $\pm$ 0.30\\ \hline
        
       $\setdist^d_{max}$  &  0.63 $\pm$ 0.56 & 0.66 $\pm$ 0.25 & 0.29 $\pm$ 0.26 & 0.38 $\pm$ 0.16 &
                              0.39 $\pm$ 0.35 & 0.47 $\pm$ 0.24 & 0.18 $\pm$ 0.16 & 0.26 $\pm$ 0.13 &
                              2.14 $\pm$ 1.87 & 2.59 $\pm$ 1.00 & 0.61 $\pm$ 0.56 & 0.94 $\pm$ 0.40 \\ \hline
       
       Time (s) & 0.01 $\pm$ 0.00 & 72.66 $\pm$ 32.57 & 0.01 $\pm$ 0.00 & 75.22 $\pm$ 35.29&
                0.01 $\pm$ 0.00 & 130.78 $\pm$ 13.945 & 0.01 $\pm$ 0.01 & 129.13 $\pm$ 134. 66 &
                0.28 $\pm$ 0.01 & 338.89 $\pm$ 11.59 & 0.27 $\pm$ 0.04 & 340.26 $\pm$ 117.52\\
       \hline
    \end{tabular}
    }
    \caption{Comparison between \texttt{DiCE} and our method (angle-based) on $50$ instances for $\beta = 0.5$ and no minimisation. We report average and standard deviation for each metric (validity excluded). 
    % Best results for each metric highlighted in bold.
    }
    \label{tab:final-results-angle-nomin}
\end{table*}

\subsection{Evaluation metrics}
\label{sec:metrics}

We evaluate results obtained using metrics that are specifically designed to assess the proximity and diversity of the explanations returned, as well as their robustness with respect to minor changes in the input to be explained. Formally, given a distance metric $dist: \dom \times \dom \rightarrow \mathbb{R}^+$, a factual input $\cinput$ and a set of diverse counterfactuals $S = \{\cinput'_1,\ldots, \cinput'_n\}$, we consider:

\begin{itemize}
    \item $k$-distance~\cite{MohammadiKBV21}, defined as:
\begin{equation}
    k\text{-distance}(\cinput, S) = \frac{1}{\lvert S \rvert} \sum_{j=1}^{\lvert S \rvert} dist(\cinput, \cinput'_j)
\end{equation}
to measure the distance of the diverse set of counterfactuals from the factual input. Low values imply lower cost of recourse.

\item $k$-diversity~\cite{MohammadiKBV21}, defined as:

\begin{equation}
    k\text{-diversity}(S) = \frac{1}{{\lvert S \rvert \choose 2}} \sum_{j=1}^{\lvert S \rvert -1} \sum_{l=j+1}^{\lvert S \rvert} dist(\cinput'_j, \cinput'_l)
\end{equation}

to measure the distance between counterfactual explanations within the set $S$. Higher values indicate more diversity.
    
    \item in order to compare the robustness of generated counterfactuals,
we use the average and maximum set distance as defined in equations
\eqref{def_set_dist_avg} and
\eqref{def_set_dist_max}
\end{itemize}

In all our experiments we use the $L_1$ and $L_2$ distances to measure $dist$, as commonly done in the literature~\cite{Wachter_17,MohammadiKBV21,DuttaLMTM22,Jiang+23}. 

\subsection{Evaluating robustness}
\label{sec:sets_exp}

This experiment is designed to show that our approach is able to generate diverse sets of explanations that are more robust than state-of-the-art algorithms. For each dataset, we select $20$ additional instances from the test set and generate sets of counterfactual explanations each following the protocol described earlier in this section. The number of diverse counterfactuals for each input is limited to a maximum of $5$ for each method so as to limit the cognitive load on the user. For better legibility, we only report results for three datasets; the full results can be found in the appendix.
% obtained for two configurations $(i)$ $\beta = 0.5$ and $\gamma =0.1$, and $(ii)$ $\beta = 0.5$ with no minimisation, under both distance- and angle-based diversity filters.  

Table~\ref{tab:final-results-angle-min} reports the results obtained for the overall best  parameterisation $\beta= 0.5$ and $\gamma = 0.1$ under angle-based filtering. As we can observe, our approach consistently outperforms \texttt{DiCE} on several of the metrics considered across all the datasets. In particular, our explanations exhibit a higher degree of robustness compared to \texttt{DiCE}'s. Indeed, the distance between the two sets of counterfactuals generated by our algorithm for the original input and its perturbed version is always smaller than that of \texttt{DiCE}, demonstrating that our approach is successful at improving the robustness of the explanations it generates. As far as diversity is concerned, the results produced by the two appraoches are comparable for \textit{diabetes} and \textit{no2}. \texttt{DiCE} generates more diverse explanations for \textit{credit} and \textit{spam}, whereas our appraoch dominates in the \textit{news} dataset. Overall we can observe that when \texttt{DiCE} achieves better diversity, it often sacrifices proximity; our approach instead always obtains better proximity, revealing a possible tension between the two metrics. Indeed, we hypothesise the diversity of our counterfactuals is affected by the minimisation of Step 4, which brings counterfactuals closer together thus leading to a decrease in $k$-diversity. Finally, we note that the time taken by \texttt{DiCE} to generate solutions is always significantly larger, reaching a two order of magnitude difference in the \textit{news} dataset. 

\subsection{The effect of minimisation}
\label{sec:no_minimisation}

To test the impact that minimisation may have on the diversity of counterfactuals generated by our method, we conducted another set of experiments where no minimisation is performed. We report the results obtained in Table~\ref{tab:final-results-angle-nomin}, again using $\beta = 0.5$ and angle-based filtering for better comparison. Overall, we observe that this configuration results in higher degrees of diversity when compared to the previous experiments. This appears to confirm our intuition that minimising distance reduces the diversity of the set returned. Removing minimisation results in higher $k$-distance from the original input; however, our approach still outperforms \texttt{DiCE} across all datasets. Overall, the robustness of our counterfactual explanations does not appear to be compromised as our approach always returns explanations that are more robust than \texttt{DiCE}'s; however we observe a slight increase in both robustness-related metrics, indicating a possible connection between $k$-diversity and robustness. Finally, we note that also in this case the runtime performance of our algorithm is superior to  \texttt{DiCE}'s across all datasets considered.

\section{Conclusions}

In this paper we studied the robustness of counterfactual explanations with respect to minor changes in the input they were generated for. We discussed several limitations of current algorithms for generating counterfactual explanations and presented a novel framework to generate explanation with interesting robustness guarantees. While theoretically
interesting, the number of counterfactuals that need to be reported
can be infinite. Therefore, we introduced an approximation scheme that uses diversity to find a compact representation of the candidate 
counterfactuals and presented an empirical evaluation of the robustness
of our approximation. Our results show that the resulting method improves the state-of-the-art in generating \emph{robust} counterfactual explanations, while also showing great advantages in terms of computational performance. Future work will focus on devising tighter approximation schemes to further strengthen the robustness guarantees our framework can offer.

\bibliography{aaai24}

\begin{thebibliography}{26}
\providecommand{\natexlab}[1]{#1}

\bibitem[{Black, Wang, and Fredrikson(2022)}]{BlackWF22}
Black, E.; Wang, Z.; and Fredrikson, M. 2022.
\newblock Consistent Counterfactuals for Deep Models.
\newblock In \emph{Proceedings of the International Conference on Learning
  Representations ({ICLR}'22)}. OpenReview.net.

\bibitem[{Dua and Graff(2017)}]{Dua2019}
Dua, D.; and Graff, C. 2017.
\newblock {UCI} Machine Learning Repository.
\newblock \url{http://archive.ics.uci.edu/ml}.
\newblock Accessed: 2022-08-30.

\bibitem[{Dutta et~al.(2022)Dutta, Long, Mishra, Tilli, and
  Magazzeni}]{DuttaLMTM22}
Dutta, S.; Long, J.; Mishra, S.; Tilli, C.; and Magazzeni, D. 2022.
\newblock Robust Counterfactual Explanations for Tree-Based Ensembles.
\newblock In \emph{Proceedings of the International Conference on Machine
  Learning ({ICML}'22)}, volume 162, 5742--5756. {PMLR}.

\bibitem[{Fernandes et~al.(2015)Fernandes, Vinagre, Cortez, and
  Sernadela}]{misc_online_news_popularity_332}
Fernandes, K.; Vinagre, P.; Cortez, P.; and Sernadela, P. 2015.
\newblock {Online News Popularity}.
\newblock UCI Machine Learning Repository.
\newblock {DOI}: https://doi.org/10.24432/C5NS3V.

\bibitem[{{FICO Community}(2019)}]{HELOC}
{FICO Community}. 2019.
\newblock {Explainable Machine Learning Challenge}.
\newblock
  \url{https://community.fico.com/s/explainable-machine-learning-challenge}.

\bibitem[{Fokkema, de~Heide, and van Erven(2022)}]{fokkema2022attribution}
Fokkema, H.; de~Heide, R.; and van Erven, T. 2022.
\newblock Attribution-based Explanations that Provide Recourse Cannot be
  Robust.
\newblock \emph{arXiv preprint arXiv:2205.15834}.

\bibitem[{Guidotti et~al.(2019)Guidotti, Monreale, Ruggieri, Turini, Giannotti,
  and Pedreschi}]{GuidottiMRTGP19}
Guidotti, R.; Monreale, A.; Ruggieri, S.; Turini, F.; Giannotti, F.; and
  Pedreschi, D. 2019.
\newblock A Survey of Methods for Explaining Black Box Models.
\newblock \emph{{ACM} Comput. Surv.}, 51(5): 93:1--93:42.

\bibitem[{Hancox{-}Li(2020)}]{Hancox-Li20}
Hancox{-}Li, L. 2020.
\newblock Robustness in machine learning explanations: does it matter?
\newblock In \emph{Proceedings of the {ACM} Conference on Fairness,
  Accountability, and Transparency (FAT*'20)}, 640--647. {ACM}.

\bibitem[{Hopkins et~al.(1999)Hopkins, Reeber, Forman, and
  Suermondt}]{misc_spambase_94}
Hopkins, M.; Reeber, E.; Forman, G.; and Suermondt, J. 1999.
\newblock {Spambase}.
\newblock UCI Machine Learning Repository.
\newblock {DOI}: https://doi.org/10.24432/C53G6X.

\bibitem[{Jiang et~al.(2023)Jiang, Leofante, Rago, and Toni}]{Jiang+23}
Jiang, J.; Leofante, F.; Rago, A.; and Toni, F. 2023.
\newblock Formalising the Robustness of Counterfactual Explanations for Neural
  Networks.
\newblock In \emph{Procedings of the 37th {AAAI} Conference on Artificial
  Intelligence ({AAAI}'23)}, 14901--14909. {AAAI} Press.

\bibitem[{Karimi et~al.(2020)Karimi, Barthe, Balle, and
  Valera}]{karimi2020model}
Karimi, A.; Barthe, G.; Balle, B.; and Valera, I. 2020.
\newblock Model-Agnostic Counterfactual Explanations for Consequential
  Decisions.
\newblock In \emph{Proceedings of the 23rd International Conference on
  Artificial Intelligence and Statistics ({AISTATS}'20)}, 895--905.

\bibitem[{Karimi et~al.(2023)Karimi, Barthe, Sch{\"{o}}lkopf, and
  Valera}]{KarimiBSV23}
Karimi, A.; Barthe, G.; Sch{\"{o}}lkopf, B.; and Valera, I. 2023.
\newblock A Survey of Algorithmic Recourse: Contrastive Explanations and
  Consequential Recommendations.
\newblock \emph{{ACM} Comput. Surv.}, 55(5): 95:1--95:29.

\bibitem[{Leofante, Botoeva, and Rajani(2023)}]{LeofanteBR23}
Leofante, F.; Botoeva, E.; and Rajani, V. 2023.
\newblock Counterfactual Explanations and Model Multiplicity: a Relational
  Verification View.
\newblock In \emph{Proceedings of the 20th International Conference on
  Principles of Knowledge Representation and Reasoning ({KR}'23)}, 763--768.

\bibitem[{Leofante and Lomuscio(2023)}]{LeofanteLomuscio23a}
Leofante, F.; and Lomuscio, A. 2023.
\newblock Towards Robust Contrastive Explanations for Human-Neural Multi-Agent
  Systems.
\newblock In \emph{Proceedings of the 22nd International Conference on
  Autonomous Agents and Multiagent Systems ({AAMAS}'23)}, 2343--2345.

\bibitem[{Mohammadi et~al.(2021)Mohammadi, Karimi, Barthe, and
  Valera}]{MohammadiKBV21}
Mohammadi, K.; Karimi, A.; Barthe, G.; and Valera, I. 2021.
\newblock Scaling Guarantees for Nearest Counterfactual Explanations.
\newblock In \emph{Proceedings of the {AAAI/ACM} Conference on AI, Ethics, and
  Society ({AIES}'21 ).}, 177--187. {ACM}.

\bibitem[{Mothilal, Sharma, and Tan(2020)}]{mothilal2020explaining}
Mothilal, R.~K.; Sharma, A.; and Tan, C. 2020.
\newblock Explaining machine learning classifiers through diverse
  counterfactual explanations.
\newblock In \emph{Proceedings of the {ACM} Conference on Fairness,
  Accountability, and Transparency (FAT*'20).}, 607--617.

\bibitem[{Pawelczyk, Broelemann, and
  Kasneci(2020)}]{pawelczyk2020counterfactual}
Pawelczyk, M.; Broelemann, K.; and Kasneci, G. 2020.
\newblock On Counterfactual Explanations under Predictive Multiplicity.
\newblock In \emph{Proceedings of the 36th Conference on Uncertainty in
  Artificial Intelligence ({UAI}'20)}, volume 124 of \emph{Proceedings of
  Machine Learning Research}, 809--818. {AUAI} Press.

\bibitem[{Pawelczyk et~al.(2023)Pawelczyk, Datta, van~den Heuvel, Kasneci, and
  Lakkaraju}]{Pawelczyketal23}
Pawelczyk, M.; Datta, T.; van~den Heuvel, J.; Kasneci, G.; and Lakkaraju, H.
  2023.
\newblock Probabilistically Robust Recourse: Navigating the Trade-offs between
  Costs and Robustness in Algorithmic Recourse.
\newblock In \emph{Proceedings of the 11th International Conference on Learning
  Representations, ({ICLR}'23)}. OpenReview.net.

\bibitem[{{ProPublica}(2016)}]{COMPASS}
{ProPublica}. 2016.
\newblock {How We Analyzed the COMPAS Recidivism Algorithm }.
\newblock
  \url{https://www.propublica.org/article/how-we-analyzed-the-compas-recidivism-algorithm}.

\bibitem[{Slack et~al.(2021)Slack, Hilgard, Lakkaraju, and Singh}]{SlackHLS21}
Slack, D.; Hilgard, A.; Lakkaraju, H.; and Singh, S. 2021.
\newblock Counterfactual Explanations Can Be Manipulated.
\newblock In \emph{Advances in Neural Information Processing Systems 34
  (NeurIPS'21)}, 62--75.

\bibitem[{Smith et~al.(1988)Smith, Everhart, Dickson, Knowler, and
  Johannes}]{smith1988using}
Smith, J.~W.; Everhart, J.~E.; Dickson, W.; Knowler, W.~C.; and Johannes, R.~S.
  1988.
\newblock Using the ADAP learning algorithm to forecast the onset of diabetes
  mellitus.
\newblock In \emph{Proceedings of the annual symposium on computer application
  in medical care}, 261. American Medical Informatics Association.

\bibitem[{Stepin et~al.(2021)Stepin, Alonso, Catal{\'{a}}, and
  Pereira{-}Fari{\~{n}}a}]{StepinACP21}
Stepin, I.; Alonso, J.~M.; Catal{\'{a}}, A.; and Pereira{-}Fari{\~{n}}a, M.
  2021.
\newblock A Survey of Contrastive and Counterfactual Explanation Generation
  Methods for Explainable Artificial Intelligence.
\newblock \emph{{IEEE} Access}, 9: 11974--12001.

\bibitem[{Upadhyay, Joshi, and Lakkaraju(2021)}]{upadhyay2021towards}
Upadhyay, S.; Joshi, S.; and Lakkaraju, H. 2021.
\newblock Towards Robust and Reliable Algorithmic Recourse.
\newblock In \emph{Advances in Neural Information Processing Systems 34
  ({NeurIPS}'21)}, 16926--16937.

\bibitem[{{Van Looveren} and Klaise(2021)}]{van2021interpretable}
{Van Looveren}, A.; and Klaise, J. 2021.
\newblock Interpretable Counterfactual Explanations Guided by Prototypes.
\newblock In \emph{Proceedings of the European Conference on Machine Learning
  and Knowledge Discovery in Databases ({ECML} {PKDD}'21)}, 650--665.

\bibitem[{Vanschoren et~al.(2013)Vanschoren, van Rijn, Bischl, and
  Torgo}]{OpenML2013}
Vanschoren, J.; van Rijn, J.~N.; Bischl, B.; and Torgo, L. 2013.
\newblock OpenML: networked science in machine learning.
\newblock \emph{{SIGKDD} Explor.}, 15(2): 49--60.

\bibitem[{Wachter, Mittelstadt, and Russell(2017)}]{Wachter_17}
Wachter, S.; Mittelstadt, B.~D.; and Russell, C. 2017.
\newblock Counterfactual Explanations without Opening the Black Box: Automated
  Decisions and the {GDPR}.
\newblock \emph{Harv. JL \& Tech.}, 31: 841.

\end{thebibliography}

\iftrue
\clearpage

\section{Proofs of Technical Results}

\setcounter{lemma}{0}
\setcounter{proposition}{0}
\begin{lemma}
\begin{itemize}
    \item For all distance measures $d$, and all $\emptyset \subset S_1, S_2 \subseteq \dom$, 
    $\setdist^d_{\sum}(S_1,S_2) \leq \setdist^d_{\max}(S_1,S_2)$.
    \item If $d$ satisfies \emph{Symmetry}, $S_1 = \{\ccf_1\}$ and $S_2 = \{\ccf_2\}$ ($S_1$ and $S_2$ contain a single point),
    then $\setdist^d_{\sum}(S_1,S_2) = \setdist^d_{\max}(S_1,S_2) = d(\ccf_1,\ccf_2)$.
\end{itemize}    
\end{lemma}
\begin{proof}
1. The claim follows from observing that  
$\frac{1}{\lvert S \rvert} \sum_{\ccf \in S} \min_{\ccf' \in S'} d(\ccf, \ccf')
\leq \frac{\lvert S \rvert}{\lvert S \rvert}  \max_{\ccf \in S} \min_{\ccf' \in S'} d(\ccf, \ccf')
$ and the definitions.

2. Both measures evaluate to $\frac{1}{2} d(\ccf_1, \ccf_2) + \frac{1}{2} d(\ccf_2, \ccf_1)$, which is just
$d(\ccf_1, \ccf_2)$ by \emph{Symmetry}.
\end{proof}

\begin{lemma}
If $d$ satisfies the triangle inequality, then
for all  $\cinput_1, \cinput_2 \in \dom$
such that 
$\classifier(\cinput_1) = \classifier(\cinput_2)$,
we have $\cfd(\cinput_1) \leq d(\cinput_1, \cinput_2) + \cfd(\cinput_2)$.
\end{lemma}
\begin{proof}
 First assume that there is a strong
 counterfactual $\ccf_2$ for
 $\cinput_2$. 
 Since $\classifier(\cinput_1) = \classifier(\cinput_2)$, 
 $\ccf_2$ is a (not necessarily strong) counterfactual
 for $\cinput_1$ as well and we have
 $\cfd(\cinput_1) \leq d(\cinput_1, \ccf_2) \leq d(\cinput_1, \cinput_2)  + d(\cinput_2, \ccf_2) = d(\cinput_1, \cinput_2)  + \cfd(\cinput_2)$.

 If there is no strong counterfactual for
 $\cinput_2$, then,
 for every $\epsilon > 0$,
 we can find a (non-strong) counterfactual $\ccf_2$ 
 such that $d(\cinput_2,\cinput_2) <
 \cfd(\cinput_2) + \epsilon$.
 As before, we can then conclude that
 $\cfd(\cinput_1) \leq d(\cinput_1, \cinput_2) + \cfd(\cinput_2) + \epsilon$.
 Since $\epsilon$ can be arbitrarily small,
 we can conclude that 
 $\cfd(\cinput_1) \leq d(\cinput_1, \cinput_2) + \cfd(\cinput_2)$.
\end{proof}
\begin{proposition}
If $d$ satisfies \emph{Symmetry} and
the \emph{Triangle Inequality}, then
$E^{\epsilon}_{exh}$ is weakly $\frac{\epsilon}{2}$-robust.    
\end{proposition}
\begin{proof}
Consider two inputs $\cinput_1, \cinput_2 \in \dom$
such that $d(\cinput_1, \cinput_2) < \frac{\epsilon}{2}$.
Assume that 
$\ccf_1 \in E(\classifier, \cinput_1)$
is a strong counterfactual wrt. $\cinput_1$.
Therefore,
$d(\cinput_2, \ccf_1) \leq 
d(\cinput_2, \cinput_1) + d(\cinput_1, \ccf_1)
< \frac{\epsilon}{2} + \cfd(\cinput_1)
\leq \frac{\epsilon}{2} + (\cfd(\cinput_2) + \frac{\epsilon}{2})
= \cfd(\cinput_2) + \epsilon.
$
Hence, $\ccf_1$ is an $\epsilon$-approximate
counterfactual for $\cinput_2$ and therefore
$\cinput_{\ccf_1} \in E^{\epsilon}_{exh}(\classifier, \cinput_2)$.
\end{proof}
We explain the geometrical intuition of the
result in Figure \ref{fig:prop}.
\begin{figure}[tb]
	\centering
		\includegraphics[width=\columnwidth]{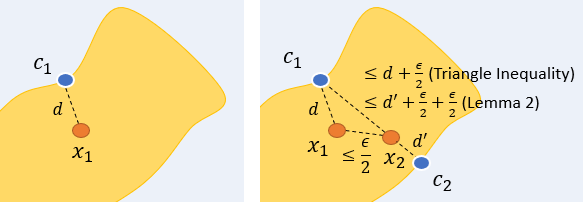}
	\caption{Illustration of the geometrical intuition of Proposition 1: As we move from $x_1$ to $x_2$ by less than
 $\frac{\epsilon}{2}$, the distance from the
 strong counterfactual $c_1$ for $x_1$ to $x_2$ cannot 
 be larger than the counterfactual distance
 of $x_1$ plus $\frac{\epsilon}{2}$ by the
 triangle inequality. Furthermore, Lemma 2 
 guarantees that the difference between the counterfactual distances of $x_1$ and $x_2$ cannot be larger than $\frac{\epsilon}{2}$. Hence,
 the distance from $c_1$ to $x_2$ cannot be
 larger than $x_2$'s counterfactual distance
 plus $\epsilon$. $E^{\epsilon}_{exh}$ will
 therefore report it (see the proof of Proposition 1 for the algebraic argument).} 
	\label{fig:prop}
\end{figure}

\begin{proposition}
\label{label_prop_safe_cf}
Suppose that $d$ satisfies \emph{Symmetry} and
the \emph{Triangle Inequality}.
If 
$\ccf \in E^{\epsilon}_{exh}(\classifier, \cinput)$
is $\delta$-safe, then for all
$\cinput' \in \dom$
such that
$\classifier(\cinput) = \classifier(\cinput')$
and
$d(\cinput, \cinput') < \frac{\delta}{2}$,
we have $\ccf \in E^{\epsilon}_{exh}(\classifier, \cinput')$.
\end{proposition}
\begin{proof}
We have
$d(\cinput', \ccf) \leq 
d(\cinput', \cinput) + d(\cinput, \ccf)
= \frac{\delta}{2} + (\cfd(\cinput) + \epsilon - \delta)
\leq
\frac{\delta}{2} + (\cfd(\cinput') + \frac{\delta}{2}) + \epsilon - \delta
= \cfd(\cinput') + \epsilon
$
and therefore $\ccf \in E^{\epsilon}_{exh}(\classifier, \cinput')$.
\end{proof}

\begin{proposition}
Consider a classification problem with $k$ features and $N$ examples.
\begin{itemize}
    \item Step 1 can be computed in time $O(N \cdot (k + \log N))$.
    \item Step 2 can be computed in time $O(N \cdot k)$.
    \item Let $M$ be the number of points remaining after step 2. Assuming that the distance can be
    computed in time $O(k)$, Step 3 can be computed in time
     $O(M^2 \cdot k)$.
     \item Let $T_{\classifier}(k)$ be the runtime function of the classifier and let 
     $D_{max}$ be the maximum distance between the reference point $\cinput$ and one of the remaining
     points. 
     Step 4 can be computed in time $O(M \cdot (k + T_{\classifier}(k)) \cdot \log_2 \frac{D_{max}}{\gamma})$.
\end{itemize}    
\end{proposition}
\begin{proof}
1. Computing the distances from the examples to the reference point $\cinput$ has cost $O(N \cdot k)$. Ordering the points according to the distance with QuickSort
has cost $O(N \cdot \log N)$.

2. Since the points have been ordered in step 1, we can just run through the array once, determine the cutoff point and copy the first points to a new array.
The overall cost is $O(N \cdot k)$.

3. The first point can just be added to a new array. For the $i$-th point, $i>1$, we have to compute the distance/angle to the previously added points. Since at most $i-1$
points have been added at this point, the cost in the $i$-th iteration is $O((i-1)\cdot k)$ and the overall cost is $O(M^2 \cdot k)$.

4. Let $D = d(\cinput, \ccf)$. Binary search updates $\cinput$ and $\ccf$ such that we halve the distance in every iteration. Hence, the distance after iteration $i$
is $\frac{D}{2^i}$. Hence, the termination criterion is reached after $\log_2 \frac{D}{\gamma}$ iterations. Computing $\cinput'$ and making the assignments can be done
in time $O(k)$. Hence, the overall cost per iteration is $O(k + T_{\classifier}(k))$ an=d the overall cost of one binary search is 
$O((k + T_{\classifier}(k)) \cdot \log_2 \frac{D}{\gamma})$. Since we have to run the search for $M$ examples, the overall runtime is obtained from this bound by multiplying
by $M$. 
\end{proof}

\section{Dataset details}

Our experimental analysis uses five datasets for binary classification tasks, namely:
\begin{itemize}
    \item the \textit{diabetes} dataset, which is used to predict whether a patient has diabetes or not based on diagnostic measurements;
    \item the \textit{no2} dataset, which is used to predict whether nitrogen dioxide levels exceed a given threshold, based on measurements related to traffic volumes and metereological conditions;
    \item the \textit{credit} dataset, which is used to predict the credit risk of a person (good or bad) based on a set of attribute describing their credit history;
    \item the \textit{spambase} is used to predict whether an email is to be considered spam or not based on selected attributes of the email;
    \item the \textit{online news popularity} dataset, referred to as \textit{news} in the following, is used to predict the popularity of online articles.
\end{itemize}

Table~\ref{tab:datasets} summarises the main features of the dataset used.

   \begin{table}[h!]
    \centering
    \resizebox{0.9\columnwidth}{!}{
    \begin{tabular}{cccc}
        \hline 
        \textbf{dataset} & \textbf{type} & \textbf{instances} & \textbf{variables} \\
        \hline
        \textit{no2} & numeric & 500 & 7 \\
         \textit{diabetes} & numeric & 768 & 8 \\
        \textit{credit} & numeric & 2000 & 10 \\
        \textit{spambase} & numeric & 4600 & 57 \\
        \textit{online news} & numeric & 39644 & 58 \\\hline

    \end{tabular}}
    \caption{Dataset details.}
    \label{tab:datasets}
\end{table}

Each dataset is split into a training set and a test set using the \texttt{train\_test\_split} method provided by the \texttt{sklearn} Python library\footnote{\url{https://scikit-learn.org/stable/modules/generated/sklearn.model_selection.train_test_split.html}}, run with default parameters. Min max scaling is applied to all datasets to ease training.

\section{Models}

Training is performed using the PyTorch library. We use batch size of $8$ and $100$ epochs for each dataset and model. Table~\ref{tab:accuracy} reports the accuracy obtained for each dataset.

 \begin{table}[h!]
    \centering
    % \resizebox{0.9\}{!}{
    \begin{tabular}{cc}
        \hline 
        \textbf{dataset} & \textbf{accuracy} \\
        \hline
        \textit{no2} &  0.60 \\
         \textit{diabetes} & 0.70 \\
        \textit{credit} &  0.93 \\
        \textit{spambase} & 0.91 \\
        \textit{online news} & 0.65 \\\hline

    \end{tabular}
    % }
    \caption{Model accuracy for each dataset.}
    \label{tab:accuracy}
\end{table}

\section{Hyperparameter tuning}

We performed hyperparameter tuning using $50$ counterfactual pairs not contained in our test set. We used $\beta \in \{0,0.1,0.2,0.3,0.4,0.5\}$ to assess the impact of our filtering criterion in Step 3 using both distance and angle-based filtering. Finally, we used $\gamma\in\{0.1,0.01,0.001\}$ to evaluate different precisions in Step 4. The overall best results were obtained using $\beta = 0.5$ and $\gamma = 0.1$, for which we report results in Tables~\ref{tab:tuning-results-angle-min}, Table~\ref{tab:tuning-results-dist-min}, Table~\ref{tab:tuning-results-angle-nomin} and Table~\ref{tab:tuning-results-dist-nomin}. Results obtained for other parameter valuations can be found at the following url: \url{https://github.com/fraleo/robust_counterfactuals_aaai24/tree/main/results}. Angle-based filtering proved to yield the best overall results when compared to distance-based filtering; we therefore decided to use angle-based filtering for the experiments reported in the main body of the paper.

\begin{table*}[t!]
    \centering
    \resizebox{1\textwidth}{!}{
    \begin{tabular}{ccccc|cccc|cccc|cccc|cccc}
        \cline{2-21}&
        \multicolumn{4}{c}{\textbf{diabetes}} &
        \multicolumn{4}{c}{\textbf{no2}} &
        \multicolumn{4}{c}{\textbf{credit}} &
        \multicolumn{4}{c}{\textbf{spambase}} &
        \multicolumn{4}{c}{\textbf{news}} \\
        \cline{2-21}
        & ours (L1) & \texttt{DiCE} (L1)  & ours (L2) & \texttt{DiCE} (L2) & ours (L1) & \texttt{DiCE} (L1) & ours (L2) & \texttt{DiCE} (L2) & ours (L1) & \texttt{DiCE} (L1) & ours (L2) & \texttt{DiCE} (L2) & ours (L1) & \texttt{DiCE} (L1) & ours (L2) & \texttt{DiCE} (L2) & ours (L1) & \texttt{DiCE} (L1) & ours (L2) & \texttt{DiCE} (L2) \\
        \hline
         validity & 50/50 & 50/50 & 50/50 & 50/50 &
                    50/50 & 50/50 & 50/50 & 50/50 & 
                    50/50 & 50/50 & 50/50 & 50/50 & 
                    50/50 & 48/50 & 50/50 & 48/50 & 
                    50/50 & 50/50 & 50/50 & 50/50 \\ \hline
         
        $k$-distance  & \textbf{1.22} & 1.90 & \textbf{0.55} & 1.09 & 
                        \textbf{0.78} & 1.28 &\textbf{ 0.37} & 0.83 & 
                        \textbf{0.78} & 1.91 &\textbf{ 0.39} & 1.18 & 
                        \textbf{0.74 }& 7.83 & \textbf{0.26} & 1.94 & 
                        \textbf{2.60} & 4.18 & \textbf{0.75} & 1.66\\ \hline
        
        $k$-diversity  & \textbf{1.39} & 1.29 & 0.63 & \textbf{0.75} & 
                         \textbf{0.96} & 0.91 & 0.45 & \textbf{0.54} & 
                         0.96 & \textbf{2.02} & 0.46 & \textbf{1.09} & 
                         0.45 & \textbf{8.62} & 0.15 &\textbf{ 2.00} & 
                         \textbf{3.36} & 2.04 &\textbf{0.94} & 0.84\\ \hline
        
        $\setdist^d_{\Sigma}$ & \textbf{0.25} & 0.33 & \textbf{0.11} & 0.18 & 
                                \textbf{0.19} & 0.27 & \textbf{0.09} & 0.16 & 
                                \textbf{0.35} & 0.78 & \textbf{0.18} & 0.42 & 
                                \textbf{0.59} & 3.73 & \textbf{0.15} & 0.93 & 
                                \textbf{0.71} & 1.89 & \textbf{0.18} & 0.67 \\ \hline
        
       $\setdist^d_{max}$  &  \textbf{0.59} & 0.64 & \textbf{0.26} & 0.36 & 
                              \textbf{0.41} & 0.47 & \textbf{0.19} & 0.28 & 
                              \textbf{0.71} & 1.35 & \textbf{0.35} & 0.73 & 
                              \textbf{0.73} & 7.12 & \textbf{0.20} & 1.69 & 
                              \textbf{1.82} & 2.90 & \textbf{0.49} & 1.06 \\ \hline
       
       Time (s) & \textbf{0.02} & 69.66 & \textbf{0.02} & 69.19 & 
                  \textbf{0.02} & 138.66 & \textbf{0.02} & 140.18 & 
                  \textbf{0.22} & 374.92 &\textbf{0.22} & 381.57 &
                  \textbf{ 0.15} & 188.95 & \textbf{0.14} & 176.75 & 
                  \textbf{0.29} & 351.52 & \textbf{0.30} & 320.53 \\
       \hline
    \end{tabular}
    }
    \caption{Comparison between our method (angle-based) and \texttt{DiCE} on $50$ instances for $\beta = 0.5$ and $\gamma = 0.1$. 
    % Best results for each metric highlighted in bold.
    }
    \label{tab:tuning-results-angle-min}
\end{table*}
\begin{table*}[t!]
    \centering
    \resizebox{1\textwidth}{!}{
    \begin{tabular}{ccccc|cccc|cccc|cccc|cccc}
        \cline{2-21}&
        \multicolumn{4}{c}{\textbf{diabetes}} &
        \multicolumn{4}{c}{\textbf{no2}} &
        \multicolumn{4}{c}{\textbf{credit}} &
        \multicolumn{4}{c}{\textbf{spambase}} &
        \multicolumn{4}{c}{\textbf{news}} \\
        \cline{2-21}
        & ours (L1) & \texttt{DiCE} (L1)  & ours (L2) & \texttt{DiCE} (L2) & ours (L1) & \texttt{DiCE} (L1) & ours (L2) & \texttt{DiCE} (L2) & ours (L1) & \texttt{DiCE} (L1) & ours (L2) & \texttt{DiCE} (L2) & ours (L1) & \texttt{DiCE} (L1) & ours (L2) & \texttt{DiCE} (L2) & ours (L1) & \texttt{DiCE} (L1) & ours (L2) & \texttt{DiCE} (L2) \\
        \hline
         validity & 50/50 & 50/50 & 50/50 & 50/50 & 
                    50/50 & 50/50 & 50/50 & 50/50 & 
                    50/50 & 50/50 & 50/50 & 50/50 & 
                    50/50 & 48/50 & 50/50 & 48/50 & 
                    50/50 & 50/50 & 50/50 & 50/50 \\ \hline
         
        $k$-distance  & \textbf{1.13} & 1.90 & \textbf{0.57} & 1.09 & 
                        \textbf{0.71} & 1.28 &\textbf{ 0.38} & 0.83 & 
                        \textbf{0.63} & 1.91 &\textbf{ 0.42} & 1.18 & 
                        \textbf{0.74 }& 7.83 & \textbf{0.26} & 1.94 & 
                        \textbf{2.32} & 4.18 & \textbf{0.82} & 1.66\\ \hline
        
        $k$-diversity  & 0.93 & \textbf{1.29} & 0.63 & \textbf{0.75} & 
                         0.71 & \textbf{0.91} & 0.47 & \textbf{0.54} & 
                         0.53 & \textbf{2.02} & 0.50 & \textbf{1.09} & 
                         0.55 & \textbf{8.62} & 0.13 &\textbf{ 2.00} & 
                         \textbf{2.48} & 2.04 &\textbf{ 1.03} & 0.84\\ \hline
        
        $\setdist^d_{\Sigma}$ & \textbf{0.12} & 0.33 & \textbf{0.13} & 0.18 & 
                                \textbf{0.13} & 0.27 & \textbf{0.11} & 0.16 & 
                                \textbf{0.10} & 0.78 & \textbf{0.22} & 0.42 & 
                                \textbf{0.36} & 3.73 & \textbf{0.17} & 0.93 & 
                                \textbf{0.42} & 1.89 & \textbf{0.63} & 0.67 \\ \hline
        
       $\setdist^d_{max}$  &  \textbf{0.44} & 0.64 & \textbf{0.26} & 0.36 & 
                              \textbf{0.35} & 0.47 & \textbf{0.23} & 0.28 & 
                              \textbf{0.33} & 1.35 & \textbf{0.37} & 0.73 & 
                              \textbf{0.65} & 7.12 & \textbf{0.21} & 1.69 & 
                              \textbf{0.93} & 2.90 & \textbf{0.23} & 1.06 \\ \hline
       
       Time (s) &   \textbf{0.03} & 69.66 & \textbf{0.02} & 69.19 & 
                    \textbf{0.02} & 138.66 & \textbf{0.02} & 140.18 & 
                    \textbf{0.19} & 374.92 &\textbf{ 0.16} & 381.57 &
                    \textbf{ 0.10} & 188.95 & \textbf{0.07} & 176.75 & 
                    \textbf{0.28} & 351.52 & \textbf{0.24} & 320.53 \\
       \hline
    \end{tabular}
    }
    \caption{Comparison between our method (distance-based) and \texttt{DiCE} on $50$ instances for $\beta = 0.5$ and $\gamma = 0.1$. 
    % Best results for each metric highlighted in bold.
    }
    \label{tab:tuning-results-dist-min}
\end{table*}
\begin{table*}[t!]
    \centering
    \resizebox{1\textwidth}{!}{
    \begin{tabular}{ccccc|cccc|cccc|cccc|cccc}
        \cline{2-21}&
        \multicolumn{4}{c}{\textbf{diabetes}} &
        \multicolumn{4}{c}{\textbf{no2}} &
        \multicolumn{4}{c}{\textbf{credit}} &
        \multicolumn{4}{c}{\textbf{spambase}} &
        \multicolumn{4}{c}{\textbf{news}} \\
        \cline{2-21}
        & ours (L1) & \texttt{DiCE} (L1)  & ours (L2) & \texttt{DiCE} (L2) & ours (L1) & \texttt{DiCE} (L1) & ours (L2) & \texttt{DiCE} (L2) & ours (L1) & \texttt{DiCE} (L1) & ours (L2) & \texttt{DiCE} (L2) & ours (L1) & \texttt{DiCE} (L1) & ours (L2) & \texttt{DiCE} (L2) & ours (L1) & \texttt{DiCE} (L1) & ours (L2) & \texttt{DiCE} (L2) \\
        \hline
         validity & 50/50 & 50/50 & 50/50 & 50/50 & 
                    50/50 & 50/50 & 50/50 & 50/50 & 
                    50/50 & 50/50 & 50/50 & 50/50 & 
                    50/50 & 48/50 & 50/50 & 48/50 & 
                    50/50 & 50/50 & 50/50 & 50/50 \\ \hline
         
        $k$-distance  & \textbf{1.45} & 1.90 & \textbf{0.66} & 1.09 & 
                        \textbf{0.97} & 1.28 & \textbf{0.46} & 0.83 & 
                        \textbf{0.82} & 1.91 & \textbf{0.41} & 1.18 & 
                        \textbf{0.97} & 7.83 & \textbf{0.34} & 1.94 & 
                        \textbf{3.42} & 4.18 & \textbf{0.99} & 1.66\\ \hline
        
        $k$-diversity  & \textbf{1.69} & 1.29 & \textbf{0.77} & 0.75 & 
                        \textbf{1.18} & 0.91 & \textbf{0.55} & 0.54 & 
                        1.00 & \textbf{2.02} & 0.45 &\textbf{ 1.09} & 
                        0.65 & \textbf{8.62} & 0.23 & \textbf{2.00} & 
                        \textbf{4.32} & 2.04 & \textbf{1.23} & 0.84\\ \hline
        
        $\setdist^d_{\Sigma}$ & \textbf{ 0.27} & 0.33 & \textbf{0.12} & 0.18 &
                                \textbf{ 0.18} & 0.27 & \textbf{0.09} & 0.16 &
                                \textbf{ 0.36} & 0.78 &\textbf{0.18} & 0.42 &
                                \textbf{ 0.50} & 3.73 & \textbf{0.17} & 0.93 & 
                                \textbf{0.58} & 1.89 & \textbf{0.17} & 0.67 \\ \hline
        
       $\setdist^d_{max}$  &  0.69 & \textbf{0.64} & \textbf{0.32} & 0.36 & 
                            \textbf{0.46}& 0.47 & \textbf{0.23} & 0.28 & 
                            \textbf{0.73} & 1.35 &\textbf{0.37} & 0.73 & 
                            \textbf{0.76} & 7.12 & \textbf{0.25} & 1.69 & 
                            \textbf{2.05} & 2.90 &\textbf{ 0.60} & 1.06 \\ \hline
       
       Time (s) & \textbf{0.01} & 69.66 & \textbf{0.01} & 69.19 &
                  \textbf{ 0.01} & 138.66 & \textbf{0.05}& 140.18 & 
                  \textbf{0.20} & 374.92 & \textbf{0.20} & 381.57 & 
                  \textbf{0.14} & 188.95 & \textbf{0.14} & 176.75 & 
                  \textbf{0.28} & 351.52 & \textbf{0.28} & 320.53 \\
       \hline
    \end{tabular}
    }
    \caption{Comparison between our method (angle-based) and \texttt{DiCE} on $50$ instances for $\beta = 0.5$ and no minimisation. 
    % Best results for each metric highlighted in bold.
    }
    \label{tab:tuning-results-angle-nomin}
\end{table*}
\begin{table*}[t!]
    \centering
    \resizebox{1\textwidth}{!}{
    \begin{tabular}{ccccc|cccc|cccc|cccc|cccc}
        \cline{2-21}&
        \multicolumn{4}{c}{\textbf{diabetes}} &
        \multicolumn{4}{c}{\textbf{no2}} &
        \multicolumn{4}{c}{\textbf{credit}} &
        \multicolumn{4}{c}{\textbf{spambase}} &
        \multicolumn{4}{c}{\textbf{news}} \\
        \cline{2-21}
        & ours (L1) & \texttt{DiCE} (L1)  & ours (L2) & \texttt{DiCE} (L2) & ours (L1) & \texttt{DiCE} (L1) & ours (L2) & \texttt{DiCE} (L2) & ours (L1) & \texttt{DiCE} (L1) & ours (L2) & \texttt{DiCE} (L2) & ours (L1) & \texttt{DiCE} (L1) & ours (L2) & \texttt{DiCE} (L2) & ours (L1) & \texttt{DiCE} (L1) & ours (L2) & \texttt{DiCE} (L2) \\
        \hline
         validity & 50/50 & 50/50 & 50/50 & 50/50 & 
                    50/50 & 50/50 & 50/50 & 50/50 & 
                    50/50 & 50/50 & 50/50 & 50/50 & 
                    50/50 & 48/50 & 50/50 & 48/50 & 
                    50/50 & 50/50 & 50/50 & 50/50 \\ \hline
         
        $k$-distance  & \textbf{1.30} & 1.90 & \textbf{0.72} & 1.09 & 
                        \textbf{0.90} & 1.28 & \textbf{0.49} & 0.83 & 
                        \textbf{0.67} & 1.91 & \textbf{0.45} & 1.18 & 
                        \textbf{0.92} & 7.83 & \textbf{0.35} & 1.94 & 
                        \textbf{3.05} & 4.18 & \textbf{1.08} & 1.66\\ \hline
        
        $k$-diversity  & 1.08 & \textbf{1.29} & \textbf{0.82} & 0.75 & 
                        0.87 & \textbf{0.91} & \textbf{0.57} & 0.54 & 
                        0.56 & \textbf{2.02} & 0.52 &\textbf{ 1.09} &
                        0.66 & \textbf{8.62} & 0.21 & \textbf{2.00} & 
                        \textbf{3.15} & 2.04 & \textbf{1.35} & 0.84\\ \hline
        
        $\setdist^d_{\Sigma}$ &\textbf{ 0.10} & 0.33 & \textbf{0.15} & 0.18 &
                                \textbf{ 0.11} & 0.27 & \textbf{0.12} & 0.16 &
                                \textbf{ 0.09} & 0.78 &\textbf{ 0.25} & 0.42 &
                                \textbf{ 0.21} & 3.73 & \textbf{0.20} & 0.93 & 
                                \textbf{0.17} & 1.89 & \textbf{0.29} & 0.67 \\ \hline
        
       $\setdist^d_{max}$  &  \textbf{0.45} & 0.64 & \textbf{0.32} & 0.36 &
                                \textbf{ 0.41 }& 0.47 & \textbf{0.27} & 0.28 & 
                                \textbf{0.33} & 1.35 &\textbf{ 0.39} & 0.73 &
                                \textbf{ 0.59} & 7.12 & \textbf{0.26} & 1.69 & 
                                \textbf{0.74} & 2.90 &\textbf{ 0.83} & 1.06 \\ \hline
       
       Time (s) & \textbf{0.01} & 69.66 & \textbf{0.01} & 69.19 &
                    \textbf{ 0.01} & 138.66 & \textbf{0.01 }& 140.18 & 
                    \textbf{0.18} & 374.92 & \textbf{0.15} & 381.57 & 
                    \textbf{0.09} & 188.95 & \textbf{0.07} & 176.75 & 
                    \textbf{0.25} & 351.52 & \textbf{0.24} & 320.53 \\
       \hline
    \end{tabular}
    }
    \caption{Comparison between our method (distance-based) and \texttt{DiCE} on $50$ instances for $\beta = 0.5$ and no minimisation. 
    % Best results for each metric highlighted in bold.
    }
    \label{tab:tuning-results-dist-nomin}
\end{table*}

\section{Full results Section~\ref{sec_experiments}}

Tables~\ref{tab:full-final-results-angle-min} and~\ref{tab:full-final-results-angle-nomin} report the full results obtained for the five datasets considered in Section~\ref{sec_experiments}.

\begin{table*}[t!]
    \centering
    \resizebox{1\textwidth}{!}{
    \begin{tabular}{ccccc|cccc|cccc|cccc|cccc}
        \cline{2-21}&
        \multicolumn{4}{c}{\textbf{diabetes}} &
        \multicolumn{4}{c}{\textbf{no2}} &
        \multicolumn{4}{c}{\textbf{credit}} &
        \multicolumn{4}{c}{\textbf{spambase}} &
        \multicolumn{4}{c}{\textbf{news}} \\
        \cline{2-21}
        & ours (L1) & \texttt{DiCE} (L1)  & ours (L2) & \texttt{DiCE} (L2) & ours (L1) & \texttt{DiCE} (L1) & ours (L2) & \texttt{DiCE} (L2) & ours (L1) & \texttt{DiCE} (L1) & ours (L2) & \texttt{DiCE} (L2) & ours (L1) & \texttt{DiCE} (L1) & ours (L2) & \texttt{DiCE} (L2) & ours (L1) & \texttt{DiCE} (L1) & ours (L2) & \texttt{DiCE} (L2) \\
        \hline
         validity & 100\% & 100\% & 100\% & 100\% & 100\% & 100\% & 100\% & 100\% & 100\% & 100\% & 100\% & 100\% & 100\% & 89\& & 100\% & 89\% & 100\% & 100\% & 100\% & 100\%  \\ \hline
         
        $k$-distance  & 1.13 $\pm$ 0.43 & 1.83 $\pm$ 0.35 &0.52 $\pm$ 0.20 & 1.06 $\pm$ 0.18 & 
                        0.62 $\pm$0.23 & 1.26 $\pm$ 0.22 & 0.31 $\pm$ 0.11 & 0.80 $\pm$ 0.15 & 
                        0.85 $\pm$ 0.33 & 1.87 $\pm$0.34 & 0.41 $\pm$ 0.15 & 1.17 $\pm$ 0.18 &
                        1.12 $\pm$ 1.27 & 7.26 $\pm$ 2.73 & 0.38 $\pm$0.39 & 1.87 $\pm$ 0.57 &
                        2.70 $\pm$ 0.97 & 3.59 $\pm$ 0.78 & 0.75 $\pm$ 0.27 & 1.47 $\pm$ 0.24  \\ \hline
        
        $k$-diversity  & 1.39 $\pm$ 0.46 & 1.32 $\pm$ 0.19 & 0.63 $\pm$ 0.22 & 0.77 $\pm$ 0.10 & 
                       0.78 $\pm$ 0.28 & 0.87 $\pm$ 0.16 & 0.30 $\pm$ 0.14 & 0.50 $\pm$ 0.09 &
                       1.02 $\pm$ 0.37 & 1.98 $\pm$ 0.31 & 0.48 $\pm$ 0.16 & 1.08 $\pm$ 0.16  &
                       0.61 $\pm$ 0.50 & 8.02 $\pm$ 2.77 & 0.20 $\pm$ 0.16 & 1.92 $\pm$ 0.55 &
                       3.45 $\pm$ 1.24 & 1.91 $\pm$ 0.93 & 0.94 $\pm$ 0.35 & 0.78 $\pm$ 0.40 \\ \hline
        
        $\setdist^d_{\Sigma}$ & 0.21 $\pm$ 0.20 & 0.33 $\pm$0.1 & 0.09 $\pm$ 0.03 & 0.18 $\pm$ 0.06 &
                                0.16 $\pm$ 0.12 & 0.87 $\pm$ 0.16 & 0.07 $\pm$ 0.05 & 0.16 $\pm$ 0.07 &
                                0.35 $\pm$ 0.18 & 0.72 $\pm$ 0.21 & 0.17 $\pm$ 0.08 & 0.39 $\pm$ 0.11 &
                                0.50 $\pm$ 0.23 & 4.20 $\pm$ 1.52 & 0.14 $\pm$0.08 & 1.05 $\pm$ 0.37 &
                                0.88 $\pm$ 0.78 & 1.73 $\pm$ 0.72 & 0.21 $\pm$ 0.21 & 0.61 $\pm$ 0.30\\ \hline
        
       $\setdist^d_{max}$  &  0.51 $\pm$ 0.44 &  0.66 $\pm$ 0.25 & 0.24 $\pm$ 0.20 & 0.38 $\pm$ 0.16 &
                              0.33 $\pm$ 0.24 & 0.47 $\pm$ 0.24 & 0.16 $\pm$ 0.11 & 0.26 $\pm$ 0.13 &
                              0.76 $\pm$ 0.32 & 1.24 $\pm$ 0.31 & 0.37 $\pm$ 0.15 & 0.67 $\pm$0.18 &
                              0.73 $\pm$ 0.38 & 7.32 $\pm$ 2.37 & 0.21 $\pm$ 0.12 & 1.75 $\pm$ 0.49 &
                              1.94 $\pm$ 1.41 & 2.59 $\pm$ 1.00 & 0.52 $\pm$ 0.44 & 0.94 $\pm$ 0.40 \\ \hline
       
       Time (s) & 0.02 $\pm$ 0.00 & 72.66 $\pm$ 32.57 & 0.02 $\pm$ 0.00 & 75.22 $\pm$ 35.29 &
                0.02 $\pm$ 0.01 & 130.78 $\pm$ 13.945 & 0.02 $\pm$ 0.01 & 129.13 $\pm$ 134. 66 &
                0.22 $\pm$ 0.02 & 396.71 $\pm$ 3.72 & 0.22 $\pm$ 0.02 & 413.09 $\pm$ 23.15 &
                0.15 $\pm$ 0.01 & 203.19 $\pm$ 125.03 & 0.14 $\pm$ 0.01 & 205.58 $\pm$ 126.87 &
                0.30 $\pm$ 0.08 & 338.89 $\pm$ 11.59 & 0.30 $\pm$ 0.00 & 340.26 $\pm$ 117.52\\
       \hline
    \end{tabular}
    }
    \caption{Comparison between our method (angle-based) and \texttt{DiCE} on $50$ instances for $\beta = 0.5$ and $\gamma = 0.1$ We report average and standard deviation for each metric (validity excluded). 
    % Best results for each metric highlighted in bold.
    }
    \label{tab:full-final-results-angle-min}
\end{table*}
\begin{table*}[t!]
    \centering
    \resizebox{1\textwidth}{!}{
    \begin{tabular}{ccccc|cccc|cccc|cccc|cccc}
        \cline{2-21}&
        \multicolumn{4}{c}{\textbf{diabetes}} &
        \multicolumn{4}{c}{\textbf{no2}} &
        \multicolumn{4}{c}{\textbf{credit}} &
        \multicolumn{4}{c}{\textbf{spambase}} &
        \multicolumn{4}{c}{\textbf{news}} \\
        \cline{2-21}
        & ours (L1) & \texttt{DiCE} (L1)  & ours (L2) & \texttt{DiCE} (L2) & ours (L1) & \texttt{DiCE} (L1) & ours (L2) & \texttt{DiCE} (L2) & ours (L1) & \texttt{DiCE} (L1) & ours (L2) & \texttt{DiCE} (L2) & ours (L1) & \texttt{DiCE} (L1) & ours (L2) & \texttt{DiCE} (L2) & ours (L1) & \texttt{DiCE} (L1) & ours (L2) & \texttt{DiCE} (L2) \\
        \hline
         validity & 100\% & 100\% & 100\% & 100\% & 100\% & 100\% & 100\% & 100\% & 100\% & 100\% & 100\% & 100\% & 100\% & 89\& & 100\% & 89\% & 100\% & 100\% & 100\% & 100\%  \\ \hline
         
        $k$-distance  & 1.38 $\pm$ 0.29 & 1.83 $\pm$ 0.35 & 0.64 $\pm$ 0.14 & 1.06 $\pm$ 0.18 & 
                        0.87 $\pm$ 0.20 & 1.26 $\pm$ 0.22 & 0.43 $\pm$ 0.09 & 0.80 $\pm$ 0.15 & 
                        0.88 $\pm$ 0.33 & 1.87 $\pm$0.34 & 0.42 $\pm$ 0.15 & 1.17 $\pm$ 0.18 &
                        1.29 $\pm$ 1.23 & 7.26 $\pm$ 2.73 & 0.46 $\pm$0.38 & 1.87 $\pm$ 0.57 &
                        3.53 $\pm$ 0.90 & 3.59 $\pm$ 0.78 & 0.98 $\pm$ 0.26 & 1.47 $\pm$ 0.24  \\ \hline
        
        $k$-diversity  & 1.71 $\pm$ 0.3 & 1.32 $\pm$ 0.19 & 0.78 $\pm$ 0.15 & 0.77 $\pm$ 0.10 & 
                       1.11 $\pm$ 0.24 & 0.87 $\pm$ 0.16 & 0.55 $\pm$ 0.10 & 0.50 $\pm$ 0.09 &
                       1.06 $\pm$ 0.36 & 1.98 $\pm$ 0.31 & 0.51 $\pm$ 0.56 & 1.08 $\pm$ 0.16 &
                       0.74 $\pm$ 0.54 & 8.02 $\pm$ 2.77 & 0.26 $\pm$ 0.19 & 1.92 $\pm$ 0.55 &
                       4.35 $\pm$ 1.17 & 1.91 $\pm$ 0.93 & 1.20 $\pm$ 0.34 & 0.78 $\pm$ 0.40 \\ \hline
        
        $\setdist^d_{\Sigma}$ & 0.22 $\pm$ 0.24 & 0.33 $\pm$0.1 & 0.10 $\pm$ 0.11 & 0.18 $\pm$ 0.06 &
                                0.15 $\pm$ 0.16 & 0.87 $\pm$ 0.16 & 0.07 $\pm$ 0.08 & 0.16 $\pm$ 0.07 &
                                0.35 $\pm$ 0.18 & 0.72 $\pm$ 0.21 & 0.18 $\pm$ 0.08 & 0.39 $\pm$ 0.11 &
                                0.45 $\pm$ 0.32 & 4.20 $\pm$ 1.52 & 0.16 $\pm$ 0.10 & 1.05 $\pm$ 0.37 &
                                0.79 $\pm$ 0.97 & 1.73 $\pm$ 0.72 & 0.21 $\pm$ 0.27 & 0.61 $\pm$ 0.30\\ \hline
        
       $\setdist^d_{max}$  &  0.63 $\pm$ 0.56 & 0.66 $\pm$ 0.25 & 0.29 $\pm$ 0.26 & 0.38 $\pm$ 0.16 &
                              0.39 $\pm$ 0.35 & 0.47 $\pm$ 0.24 & 0.18 $\pm$ 0.16 & 0.26 $\pm$ 0.13 &
                              0.78 $\pm$ 0.33 & 1.24 $\pm$ 0.31 & 0.39 $\pm$ 0.15 & 0.67 $\pm$0.18 &
                              0.76 $\pm$ 0.43 & 7.32 $\pm$ 2.37 & 0.26 $\pm$ 0.15 & 1.75 $\pm$ 0.49 &
                              2.14 $\pm$ 1.87 & 2.59 $\pm$ 1.00 & 0.61 $\pm$ 0.56 & 0.94 $\pm$ 0.40 \\ \hline
       
       Time (s) & 0.01 $\pm$ 0.00 & 72.66 $\pm$ 32.57 & 0.01 $\pm$ 0.00 & 75.22 $\pm$ 35.29&
                0.01 $\pm$ 0.00 & 130.78 $\pm$ 13.945 & 0.01 $\pm$ 0.01 & 129.13 $\pm$ 134. 66 &
                0.21 $\pm$ 0.03 & 396.71 $\pm$ 3.72 & 0.21 $\pm$ 0.01 & 413.09 $\pm$ 23.15 &
                0.14 $\pm$ 0.01 & 203.19 $\pm$ 125.03 & 0.14 $\pm$ 0.01 & 205.58 $\pm$ 126.87 &
                0.28 $\pm$ 0.01 & 338.89 $\pm$ 11.59 & 0.27 $\pm$ 0.04 & 340.26 $\pm$ 117.52\\
       \hline
    \end{tabular}
    }
    \caption{Comparison between our method (angle-based) and \texttt{DiCE} on $50$ instances for $\beta = 0.5$ and no minimisation. We report average and standard deviation for each metric (validity excluded). 
    % Best results for each metric highlighted in bold.
    }
    \label{tab:full-final-results-angle-nomin}
\end{table*}

\fi

% \subsection{Parameter configuration}

\end{document}